\documentclass[twoside,11pt]{article}

\usepackage{mlhc}
\usepackage{paralist,color}
\usepackage{amsmath,bbm}
\usepackage{algorithm}
\usepackage[noend]{algorithmic}
\usepackage{caption}
\usepackage{subcaption}
\usepackage{float}
\usepackage{hyperref}
\usepackage{multirow}

\newcommand{\bW}{\mathbf{W}}
\newcommand{\bA}{\mathbf{A}}
\newcommand{\bX}{\mathbf{X}}
\newcommand{\bY}{\mathbf{Y}}

\renewcommand{\b}{\mathbb}
\newcommand{\name}{Identifiable Phenotyping}
\newcommand{\ground}{ground}
\newcommand{\sip}{MLC}

\begin{document}

\title{\name~using Constrained Non--Negative Matrix Factorization}

\author{\name Shalmali Joshi \email shalmali@utexas.edu \\
       \addr Electrical and Computer Engineering\\
       The University of Texas at Austin\\
      Austin, TX, USA 
      \AND
      \name Suriya Gunasekar \email suriya@utexas.edu \\
       \addr Electrical and Computer Engineering\\
      The University of Texas at Austin\\
       Austin, TX, USA 
      \AND
      \name David Sontag \email dsontag@cs.nyu.edu \\
       \addr Computer Science\\
      New York University\\
       NYC, NY, USA
      \AND
      \name Joydeep Ghosh \email jghosh@utexas.edu  \\
       \addr Electrical and Computer Engineering\\
       The University of Texas at Austin\\
       Austin, TX, USA } %

\maketitle

\begin{abstract}
This work proposes a new algorithm for automated and simultaneous phenotyping  of multiple co--occurring medical conditions, also referred to as comorbidities, using clinical notes from electronic health records (EHRs). A latent factor estimation technique, non-negative matrix factorization (NMF), is augmented with domain constraints from weak supervision to obtain sparse latent factors that are \emph{\ground ed} to a fixed set of chronic conditions. The proposed {\ground ing} mechanism ensures a one-to-one identifiable and interpretable mapping between the latent factors and the target comorbidities. Qualitative assessment of the empirical results by clinical experts show that the proposed model learns  clinically interpretable phenotypes which are also shown to have competitive  performance on $30$ day mortality prediction task. The proposed method can be readily adapted to any non-negative EHR data across various healthcare institutions.
\end{abstract}
\section{Introduction}\label{sec:intro}
%Phenotype definitions in the context of Electronic health records (EHRs) or 
Reliably querying for patients with specific medical conditions across multiple organizations facilitates many large scale healthcare applications such as cohort selection, multi-site clinical trials, epidemiology studies etc.~\citep{richesson2013electronic,hripcsak2013next,pathak2013electronic}.
However, raw EHR data collected across diverse populations and multiple care-givers can be extremely high dimensional, unstructured, heterogeneous, and noisy. Manually querying such data is a formidable challenge for healthcare professionals. 

\textit{EHR driven phenotypes} are concise representations of medical concepts composed of clinical features, conditions, and other observable traits  facilitating accurate querying of individuals %with a certain medical condition 
from EHRs \citep{Collaboratory:Mr-x2zyZ}. Efforts like eMerge Network\footnote{\url{http://emerge.mc.vanderbilt.edu/}}, PheKB\footnote{\url{http://phekb.org/}} are well known examples of EHR driven phenotyping. Traditionally used rule--based composing methods %involving iterative and collaborative effort of medical and IT professionals \citep{hripcsak2013next,Newton:2013fx} 
for phenotyping require substantial time and expert knowledge and have little scope for exploratory analyses. This  motivates automated EHR driven phenotyping using machine learning with limited expert intervention. %Existing machine learning tools for phenotyping use standard EHR data features such as laboratory measurements, diagnosis codes (ICD-$9$ or ICD-$10$\footnote{\url{www.cdc.gov/nchs/icd/}}), procedure codes, drug history~\citep{schmiedeskamp2009use, Ho2014:KDD,kawaler2012learning} and unstructured data such as clinical notes \citep{Nguyen440,Uzuner561}. %Invariably, such automated tools pose an inherent trade--off between the accuracy of estimation, richness of EHR data, and level of human supervision.  %Additionally, unstructured data like clinical notes can be harnessed for phenotyping \citep{Nguyen440,Uzuner561} and typically combining all available information from EHRs provides superior phenotyping results~\citep{wei2015combining,peissig2012importance}. %While a full review of machine learning methods in the context of phenotyping is beyond the scope of this paper, a brief review and comparison of relevant existing work is included in Section~\ref{sec:discussion}.

We propose a weakly supervised model for jointly phenotyping $30$ co--occurring conditions (comorbidities) observed in {intensive care unit (ICU)} patients. \emph{Comorbidities} are a set of co-occurring conditions in a patient at the time of admission that are not directly related to the primary diagnosis for hospitalization~\citep{elixhauser1998comorbidity}. Phenotypes for the $30$ comorbidities listed in Table~\ref{tab:conds1} are derived using text-based features from clinical notes in a publicly accessible MIMIC-III EHR database \citep{saeed2011multiparameter}. %We use a latent factor model that views phenotypes as a small number of latent variables generating the observed data \citep{hripcsak2013next,Ho2014:JBI, Ho2014:BIH, Ho2014:KDD, halpern2014using, halpern2016electronic}. 
We present a novel \textit{constrained non--negative matrix factorization (CNMF)} for the EHR matrix that aligns the factors with target comorbidities yielding sparse, interpretable, and identifiable phenotypes. %where 
%the EHR data matrix $\bX$ is approximated as a product of two low dimensional non--negative factor matrices as $\bX \approx \bA\bW$. We will henceforth refer to the factors as the \textit{phenotype factor matrix $\bA$} and the \textit{patient loadings matrix $\bW$}. The columns of $\bA$ are treated as candidates for phenotype definitions for the chronic conditions of interest. The 
%a basic non-negative matrix factorization (NMF)~\citep{lee1999learning} model is adapted for effective phenotyping by incorporating various domain specific constraints. %Consider a non negative low rank factorization of  EHR data of patients represented as columns of a matrix $\bX$ approximated as a product of two low dimensional non--negative factor matrices as $\bX \approx \bA\bW$. We will henceforth refer to the factors as the \textit{phenotype factor matrix $\bA$} and the \textit{patient loadings matrix $\bW$}. The columns of $\bA$ are treated as candidates for phenotype definitions for the chronic conditions of interest.

The following aspects of our model distinguish our work from prior efforts:
\begin{asparaenum}
\item \textbf{Identifiability: } A key shortcoming of standard unsupervised latent factor models such as %simple probabilistic and non--probabilistic models such as %Probabilistic Latent Semantic Indexing (PLSI) \citep{hofmann1999probabilistic}, 
  %Principal Component analysis (PCA), %Independent Component Analysis (ICA), 
NMF~\citep{lee2001algorithms} and Latent Dirichlet Allocation (LDA) \citep{blei2003latent} for phenotyping is that, the estimated latent factors learnt are interchangeable and \textit{unidentifiable} as phenotypes for specific conditions of interest. %For example, in low rank NMF, the rank--$1$ factors are invariant to permutations, and thus a factor component cannot be readily interpreted as a phenotype for a particular condition of interest.
%the columns of phenotype factor matrix $\bA$ are invariant to permutations (with appropriate permutation on rows of $\bW$), and thus column(s) of $\bA$ cannot be readily interpreted as specific phenotype(s) of interest.  %On the other hand, for fully supervised models like support vector machines \citep{carroll2011naive}, highly accurate supervision is expensive to obtain for large patient populations.
We tackle identifiability by incorporating  weak (noisy) but inexpensive supervision as constraints our framework. %for latent factor estimation. 
Specifically, we obtain weak supervision for the target conditions in Table \ref{tab:conds1} using the Elixhauser Comorbidity Index (ECI) \citep{elixhauser1998comorbidity} computed solely from patient administrative data (without human intervention). %We note that such measures provide a potentially noisy diagnosis of these conditions and using vanilla supervised classification methods might lead to higher bias (as demonstrated in our empirical study). %Specifically, we use the Elixhauser Comorbidity Index (ECI) \citep{elixhauser1998comorbidity} which uses ICD-$9$-CM billing code and drug code information to derive weak diagnoses for $30$ chronic conditions (Table~\ref{tab:conds1}). 
We then \ground ~the latent factors to have a one-to-one mapping with conditions of interest by incorporating the comorbidities predicted by ECI as \emph{support constraints} on the patient loadings along the latent factors. %In our experiments (Section~\ref{sec:exp-phenotyping}), we observe that including these support constraints generate highly relevant phenotypes for target chronic conditions.
 %We emphasize that our NMF algorithm is not fully unsupervised and the weak supervision from Elixhauser Comorbiditiy Index is generic tool that can be readily used on patients' EHR data without further human effort. %Other significant short-comings of latent factor models in detail in Section~\ref{sec:discussion}.
\item \textbf{Simultaneous modeling of comorbidities: }  %EHR driven phenotypes are typically defined in the context of 
ICU patients studied in this paper are frequently afflicted with multiple co--occurring conditions besides the primary cause for admission. %In ICU patient population, containing a preponderance of patients suffering from chronic diseases in advanced stages, patients tend to be afflicted with  multiple co--occurring chronic conditions triggered from a primary condition, also known as \textit{comorbidities}. 
In the proposed NMF model, phenotypes for such co--occurring  conditions jointly modeled to capture the resulting correlations. %rather than deriving individual phenotypes independently. 
%In the proposed model, $30$ such comorbidities are jointly modeled as components in an NMF. %Additionally, a feature bias factor is included (Section~\ref{sec:model}) to absorb the components of the EHR matrix that do not correspond to the chronic conditions of interest.
\item \textbf{Interpretability: }  %Thus, it is desirable that inferences from this stage be clinically interpretable  by the healthcare provider. %further extensively tested by domain experts before final conclusions are derived. Thus, interpretability of the phenotypes obtained by machine learning methods is critical to wider applicability of phenotyping algorithms in practice. 
For wider applicability of EHR driven phenotyping for advance clinical decision making, it is desirable that these phenotype definitions be clinically interpretable and represented as a concise set of rules. We consider the sparsity in the representations as a proxy for interpretability %. Thus, %in addition to the non--negativity and support constraints, 
and explicitly encourage conciseness of phenotypes using tuneable sparsity--inducing soft constraints.% in the optimization framework. %The strength of the sparsity--inducing penalty can be tuned to a desired level of trade--off between interpretability and  accuracy. 
%\item \textbf{Robustness: }Finally, since the algorithm proposed in this paper, as well as  algorithms for many latent factor estimation models, are  non--convex in nature, the results obtained tend to depend on parameter  initialization. We observe that in the presence of a dense bias factor to absorb spurious signals in EHR, the sparsity and support constraints facilitate consistent phenotype estimates. We include bias a component in the approximation in order to capture redundant or non-discriminative terms, further facilitating stability. We empirically evaluate the robustness of phenotype factor estimated from our model for to various random initializations in Section~\ref{sec:exp-robustness}. 
\end{asparaenum}
%Many existing work adapt latent factor estimation techniques developed in machine learning to automate and significantly speed up the phenotype extraction process.

We evaluate the effectiveness of the proposed method towards interpretability, clinical relevance, and prediction performance on EHR data from MIMIC-III. Although we focus on ICU patients using clinical notes, the proposed model and algorithm are general and can be applied on any non-negative EHR data from any population group.
%Our contributions can be summarized as follows:
%\begin{compactenum}
%\item We propose a scalable and efficient algorithm for EHR driven phenotyping for multiple chronic conditions jointly. 
%\item We show that our novel \grounding~ technique allows to ground the latent factors to chronic conditions of interest. When combined with sparsity inducing penalties on the phenotype estimates, we empirically show that our latent factors are identifiable. 
%\item Extensive experiments on EHR data from MIMIC-III  demonstrate the effectiveness of our proposed method towards various qualitative and quantitative evaluation metrics including interpretability, robustness, and prediction performance.
%\end{compactenum}
%In order to jointly extract clinically relevant phenotypes for the $30$ chronic conditions simultaneously, we use the following data extraction procedure on clinical notes available in the MIMIC-III dataset\footnote{\url{https://mimic.physionet.org/}} \citep{saeed2011multiparameter}.

%\input{data}

\section{Data Extraction}\label{sec:data}
%To discover phenotypes for multiple chronic conditions, we use clinical notes available in the MIMIC-III dataset\footnote{\url{https://mimic.physionet.org/}} \citep{saeed2011multiparameter}.
The MIMIC-III dataset consists of de-identified EHRs for $\sim38,000$ adult ICU patients at the Beth Isreal Deaconess Medical Center, Boston, Massachusetts from $2001$--$2012$. For all ICU stays within each admission, clinical notes including nursing progress reports, physician notes, discharge summaries, ECG, etc. are available. We analyze patients who have stayed in the ICU for at least $48$ hours ($\sim17000$ patients). We derive phenotypes using clinical notes collected within the first $48$ hours of patients' ICU stay to evaluate the quality of phenotypes when limited patient data is available. Further, we evaluate the phenotypes on a $30$ day mortality prediction problem. To avoid obvious indicators of mortality and comorbidities, apart from restricting to 48 hour data, we exclude discharge summaries as they explicitly mention patient outcomes (including mortality).
%We only consider patients who have lived in the hospital for at least 48 hours. Clinical notes over the first 48 hours of the ICU stay are used for analyses. However, we excluded discharge summaries since they explicitly mention patient outcomes including mortality. %We excluded patient populations with a cumulative length of stay less than $2$ days or more than $100$ days as well as those with fewer than $3$ notes or more than $150$ notes. 
%The final patient population consist of $\sim17,000$ patients.
%An aggregate clinical note is obtained per patient that combines all available notes over all admissions. 
%The billing code (ICD-$9$) and HCFA drug codes from the hospital admission corresponding to the first 48 hour stay are used to estimate the Elixhauser Comorbidity Index (ECI)~\citep{elixhauser1998comorbidity} for the $30$ chronic conditions listed in Table~\ref{tab:conds1}. The comorbiditiy index is used as weak supervision in the proposed algorithm. Note that our model is not constrained to using clinical notes and is easily generalized to any non-negative feature representation. 
\begin{table}[t]
\centering
\caption{Target comorbidities}\label{tab:conds1}
\resizebox{\textwidth}{!}{
\begin{tabular}{lllll}
\hline
%\hline
Congestive Heart Failure & Cardiac Arrhythmias & Valvular Disease   & Pulmonary Circulation Disorder &  Peripheral Vascular  Disorder  \\
Hypertension             & Paralysis          & Other Neurological Disorders &   Chronic Pulmonary Diseases    & Diabetes Uncomplicated \\ 
Diabetes Complicated     & Hypothyroidism  & Renal Failure      & Liver Disease   (excluding bleeding)      & Peptic Ulcer \\
AIDS &  Lymphoma           & Metastatic Cancer  & Solid Tumor          (without metastasis) & Rheumatoid Arthritis \\
 Coagulopathy             & Obesity            & Weight loss        & Fluid Electrolyte Disorder   &  Blood Loss Anemia    \\
 Deficiency Anemia       & Alcohol abuse      & Drug abuse      &  Psychoses             & Depression    \\
\hline
\end{tabular}}
\end{table}
\begin{asparaenum}
\item {\bf{Clinically relevant bag-of-words features:}}\label{sec:data-vocab}
%\paragraph{Bag-of-Words Feature Extraction}\label{sec:data-vocab}
 Aggregated clinical notes from all sources are represented as a single \emph{bag-of-words} features. To enhance clinical relevance, we create a custom vocabulary containing clinical terms from two sources (a) the Systematized Nomenclature of Medicine-Clinical Terms (SNOMED CT), and (b) the level-0 terms provided by the Unified Medical Language System (UMLS), consolidated into a standard vocabulary format using Metamorphosys --- an application provided by UMLS for custom vocabulary creation.\footnote{See \url{https://www.nlm.nih.gov/healthit/snomedct/} and \url{https://www.nlm.nih.gov/research/umls/}} To extract clinical terms from the raw text, the notes were tagged for chunking using a conditional random field tagger\footnote{\url{https://taku910.github.io/crfpp/}}. The tags are looked up against the custom vocabulary (generated from Metamorphosys) to obtain the \emph{bag-of-words} representation. Our final vocabulary has $\sim$3600 clinical terms.
\item {\bf{Computable weak diagnosis:}}\label{sec:data-supp}
We incorporate domain constraints from weak supervision to \ground ~the latent factors to have a one-to-one mapping with the conditions of interest. In the model described in Section~\ref{sec:model}, this is enforced by constraining the non-zero entries on patient loading along the latent factors using a weak diagnosis for comorbidities. The weak diagnoses of target comorbidities in Table~\ref{tab:conds1} are obtained using  ECI\footnote{\url{https://git.io/v6e7q}}, computed solely from patient administrative data without human annotation. We refer to this index as \textit{weak diagnoses} as it is not a physician's exact diagnosis and is subject to noise and misspecification. %In \eqref{eq:nmf}, constraints on the patient loadings matrix ensure that a patients' loadings are non-zero if and only if they are diagnosed with the corresponding chronic conditions.
%As the billing and drug code information is indirectly incorporated into the model through the ECI, we do not use this information in our feature set. 
Note that ECI ignores diagnoses code related to the primary diagnoses of admission. Thus, ECI models presence and absence of conditions other than the primary reason for admission (comorbidities). The phenotype candidates from the proposed model can be considered as concise representations of such comorbidities.
\end{asparaenum}

\section{Identifiable High--Throughput Phenotyping}\label{sec:setup}
%\subsection{Model Formulation}
The notation used in the paper are enumerated in Table~\ref{tab:notation}. 
\begin{table}[t]
\small
\centering
\begin{tabular}{lp{11cm}}
\hline
Notation                         & Description                                                                   \\ \hline
$[m]$ for integer $m$ & Set of indices $[m]=\{1,2,\dots,m\}$. \\
$\Delta^{d-1}$ & Simplex in dimension $d$, ${\Delta^{d-1}=\{x\in\mathbb{R}^d_+:\sum x_i=1\}}$. \\
$x^{(j)} $     & Column $j$  of a matrix $\bX$.  \\
$\text{supp}(x)$ & Support of a vector $x$, $\text{supp}(x)=\{i:x_i\neq 0\}$. \\
\bf Observations &\\
$N$, $d$                             & Number of patients  ($\sim17000$) and  features  ($\sim3600$), respectively.   \\
$\bX\in\mathbb{R}_+^{d\times N}$ & EHR matrix from MIMIC III: Clinically relevant bag-of-words features from notes in first $48$ hours of ICU stay for $N$ patients. \\
$k=1,2,\ldots,K$               & Indices for  $K=30$ comorbidities in Table \ref{tab:conds1}. \\
$C_j\subseteq [K]$ for $j\in[N]$        & Set of comorbidities  patient $j$ is diagnosed with using  ECI . \\
\bf Factor matrices &\\
$\tilde{\bW} \in [0,1]^{K\times N}$ & Estimate of \textit{patients' risk} for the $K$ conditions.\\%: $\bW^*$ is unknown ground truth, and $\tilde{\bW}$ is the estimate form Algorithm~\ref{alg:alg1}\\
$\tilde{\bA} \in \mathbb{R}_+^{d\times K}$, $\tilde{b}\in\mathbb{R}_+^{d}$ & Estimate of \textit{phenotype factor matrix} and \textit{feature bias vector}. \\%The columns represent phenotype definitions of $K$ chronic conditions\\%,   unknown ground truth, and  estimate form Algorithm~\ref{alg:alg1}, respectively\\
%$\tilde{b}\in\mathbb{R}_+^{d}$ & Estimated feature bias vector. \\
\hline
\end{tabular}
\caption{Notation used in the paper}
\label{tab:notation}
\end{table}
In summary, for each patient $j\in[N]$, 
\begin{inparaenum}[(a)]
\item the {bag-of-words} features from clinical notes is represented as column $x^{(j)}$ of EHR matrix $\bX\in\mathbb{R}_{+}^{d \times N}$, and 
\item  the list of comorbidities diagnosed  using ECI is denoted as $C_j\subseteq [K]$. %, where $K=1,2,\ldots, K$ index the $K=30$ chronic conditions  in Table~\ref{tab:conds1}. 
\end{inparaenum}
% \usepackage{booktabs}
 Let an unknown  $\bW^*\in[0,1]^{K\times N}$ represent the risk of $N$ patients for $K$ comorbidities of interest; each entry $w^*_{kj}$ lies in the interval $[0,1]$, with $0$ and $1$ indicating no-risk and maximum-risk, respectively, of patient $j$ being afflicted with condition $k$. If $C_j^*\subseteq[K]$ denotes an accurate diagnosis  for patient  $j$, then $w^{*(j)}$ satisfies $\text{supp}(w^{*(j)})\subseteq C_j^*$.%, i.e., $\forall k\notin C_j, w^*_{kj} =0$.% we  use the following definition of EHR driven phenotype.
\begin{definition}[EHR driven phenotype]\label{def:phenotype} \normalfont  \textit{EHR driven phenotypes} for  $K$ co--occurring  conditions are a set of vectors $\{a^{*(k)}\in\mathbb{R}^d_+: k\in[K]\}$, such that for a patient $j$ afflicted with conditions $C^*_j\subseteq[K]$,
\begin{equation}
\b{E}[x^{(j)}|w^{*(j)}] = \textstyle \sum_{k\in C^*_j}\;w^*_{kj} a^{*(k)} + b^*,
\end{equation}
where $b^*$ is a  bias representing the feature component observed  independent of the $K$ target conditions. $\bA^*\in\mathbb{R}^{d\times K}$ with $a^{*(k)}$ as columns is referred as  the \textit{phenotype factor matrix}. %Thus,   
%$$\b{E}[\bX|\bW^*]=\bA^*\bW^*+b^*\mathbbm{1}^\top,$$
%where $\mathbbm{1}$ denotes a vector of all ones (of appropriate dimension).
\end{definition}
% and without loss of generality (WLOG), let columns of $\bA^*$ be  normalized to unit $\ell_1$ norm.
% Explicitly modeling a feature bias stems from the following clinical motivation. Consider a physiological trait like temperature, which is not discriminative of chronic conditions of interest but is a typically measured trait and likely to occur in clinical notes frequently. Such terms are therefore better captured by an independent bias component. Note that these features may not necessarily be as frequent as stopwords to be removed using simple rule based dynamic stop-word removal and hence automatically learned during training.
%\paragraph{Feature Bias} 
Note that we explicitly model a feature bias $b^*$ to capture frequently occurring terms that are not discriminative of the target  conditions, e.g., temperature, pain, etc.%These features may not be removed using simple rule based stop-word removal and hence learned during training.

\paragraph{Cost Function} The bag-of-words features are represented as counts in the EHR matrix $\bX$. We consider a factorized approximation of $\bX$ parametrized by matrices $\bA \in \mathbb{R}_{+}^{d \times K}$, $\bW \in \mathbb{R}_{+}^{K \times N}$ and $b\in \mathbb{R}_{+}^{d}$ as $\bY=\bA\bW+b\mathbbm{1}^\top$, where $\mathbbm{1}$ denotes a vector of all ones of appropriate dimension. The approximation error of the estimate is measured using the $I$--divergence defined as follows:  
\begin{equation}
\textstyle\mathcal{D}(\bX,\bY)=\sum_{ij}{y_{ij}}-{x_{ij}}-x_{ij}\log{\frac{y_{ij}}{x_{ij}}}.\end{equation}
Minimizing the $I$--divergence is equivalent to maximum likelihood estimation under a Poisson distributional assumption on individual entries of the EHR matrix parameterized by $\bY=\bA\bW+b\mathbbm{1}^\top$~\citep{banerjee2005clustering}.%\footnote{I-divergence is a member of the class of Bregman Divergences~\citep{banerjee2005clustering}. This class of divergences is attractive because of properties such as convexity and associations to the popular exponential family distributions. The model can be generalized to a rich set of datatypes and probabilistic models by an appropriate choice of the divergence function within this class.}. 

%We consider the following factorization of the EHR matrix:
%\begin{equation}
%\textstyle\bX\overset{\mathcal{D}}{\sim}\bA\bW+b\mathbbm{1}^\top,
%\label{eq:approx}
%\end{equation}
%where $\bX\overset{\mathcal{D}}{\sim}\bY$ denotes an approximation obtained as $\text{argmin}_{\bY\in\mathcal{Y}}\mathcal{D}(\bX,\bY)$, for some constraint set $\mathcal{Y}$ and $\mathbbm{1}$ denotes a vector of all ones of appropriate dimension.
\paragraph{Phenotypes} For the $K$ comorbidities,  columns of $\bA$, $\{a^{(k)}\}_{k\in[K]}$ are  proposed as candidate  phenotypes derived from the EHR $\bX$, i.e.  approximations to $\{a^{*(k)}\}_{k\in[K]}$.
\paragraph{Constraints} %In a basic NMF, divergence minimization under just the non--negativity constraints. This vanilla NMF may not provide an identifiable or sufficiently interpretable (sparse) factorization. 
The following constraints are incorporated in learning $\bA$ and $\bW$.
\begin{asparaenum}
\item \textit{Support Constraints:} The non-negative rank--$K$ factorization of $\bX$ is `\ground ed' to $K$ target comorbidities by constraining the support of risk $w^{(j)}$ corresponding to patient $j$ using weak diagnosis $C_j$ from ECI  %If $C_j$ are accurate,  then this constraint follows from the definition of phenotypes. 
as an approximation of the conditions in Definition~\ref{def:phenotype}.   %EHR for a patient $j$ is approximated by a non--negative linear combination of phenotype signatures of the diagnosed chronic conditions only:
%\begin{equation}
%x^{(j)} \overset{\mathcal{D}}{\sim} \sum_{k \in C_j} w_{kj}a^{(k)} +b_j.
%\end{equation}
\item \textit{Sparsity Constraints:}  %As noted earlier, in many applications,  results from phenotyping are eventually consumed by domain experts for final decision making. Thus, it is desirable that the phenotype representations be easily interpretable for human experts. 
%Sparsity of the phenotypes $\{a^{(k)}\}_{k\in[K]}$ is used as a measure of clinical interpretability. 
Scaled simplex constraints are imposed on the columns of $\bA$ with a tuneable parameter $\lambda>0$ to encourage sparsity of phenotypes. Restricting the patient loadings matrix as $\bW \in [0,1]^{K \times N}$ not only allows to interpret the loadings as the patients' risk, but also makes simplex constraints effective in a bilinear optimization. %in addition to simplex constraints, 
%we require a restriction on some norm of the $\bW$ matrix. Without loss of generality, in the optimization, we restrict the patient loadings matrix as $\bW \in [0,1]^{K \times N}$. This constraint has the added advantage that for every patient $j$, the loadings can be interpreted as the propensity/risk of patient $j$ to suffer from each condition.
\end{asparaenum}

Simultaneous phenotyping of comorbidities using constrained NMF is posed as follows:
\begin{equation}\label{eq:opt}
\begin{aligned}
\tilde{\bA}, \tilde{\bW}, \tilde{b} = &{\text{argmin}}_{\bA\ge 0,\bW\ge 0, b\ge 0}&& \mathcal{D}( \bX, \bA\bW+b\mathbbm{1}^\top) & \\ %+\lambda \|\bW\|_F^2& \\
&\text{ s.t. }&& \text{supp}(w^{(j)}) = C_j  \ \forall j \in [N], \; \bW \in [0,1]^{K \times N},&\\
&&& a^{(k)}\in \lambda \Delta^{d-1} \ \forall j\in[K],
\end{aligned}
\end{equation}
%where recall  that $\text{supp}(w^{(j)})=C_j\Rightarrow \forall k\notin C_j, w_{kj} =0$, $\Delta^{d-1}$ is the scaled simplex in $\b{R}_{+}^d$.%, column $a^{(k)}$ of the estimated matrix $\tilde{\bA}$ is a candidate phenotype for the chronic condition $k \in [K]$, and $w^{(j)}_k$  (or $w_{kj}$) is the estimated risk of condition $k$ for patient $j$.
%$\mathcal{D}$ in \eqref{eq:nmf} is assumed to belong to a class of 
The optimization in \eqref{eq:opt} is convex in either factor with the other factor fixed. It is solved using alternating minimization with projected gradient descent~\citep{parikh2014proximal,lin2007projected}. See complete algorithm in Algorithm~\ref{alg:alg1}. The proposed model in general can incorporate any weak diagnosis of medical conditions. 
In this paper, we note that, since we use ECI, the results are not representative of the primary diagnoses at admission. 
\begin{algorithm}[t]
\caption{Phenotyping using constrained NMF. \\Input:  $\bX$, $\{C_j: j \in [N]\}$ and paramter $\lambda$. Initialization: $\bA_{(0)}, b_{(0)}$. }\label{alg:alg1}
\begin{algorithmic}
%\STATE{Input:}
\WHILE{Not converged}
\STATE{$\bW_{(t)} \leftarrow \arg\min_{\bW} \mathcal{D}(\bX,\bA_{(t-1)}\bW+b_{(t-1)}\mathbbm{1}^\top)\text{ s.t. }\bW \in [0,1]^{K \times N}, \text{supp}(w^{j}) = C_j, \ \forall j$}%+\lambda\|\bW\|_F^2$}
\STATE{$\bA_{(t)}, b_{(t)}\leftarrow \arg\min_{\bA, b\ge 0} \mathcal{D}(\bX,\bA\bW_{(t)}+b\mathbbm{1}^\top) \text{ s.t. } a_j^{(k)} \in \lambda \Delta^{d-1}, \forall k$}
\ENDWHILE
%\STATE{Return $\bA_{(t)}, \bW_{(t)}$}
\vspace{-2pt}
\end{algorithmic}
\end{algorithm}
%Let $P_k$ be the set of phenotypes for condition $k$. Then from the estimated phenotype matrix $\bA$, candidate phenotypes $P_k$ are given by:
%\begin{equation}
%P_{k} = \{j : a_{j}^{(k)} > 0 , j \in [d] \}
%\end{equation}
\label{sec:model}
\section{Empirical Evaluation}\label{sec:expts}
The estimated phenotypes are evaluated on various metrics. We denote the model learned using Algorithm~\ref{alg:alg1} with a given parameter $\lambda>0$ as $\lambda$--CNMF. The following baselines are used for comparison:
\begin{asparaenum}[1.]
\item \textbf{Labeled LDA (LLDA):} LLDA~\citep{ramage2009labeled} is the supervised counterpart of LDA, a probabilistic model to estimate topic distribution of a corpus. It assumes that word counts of documents arise from multinomial distributions. It incorporates supervision on topics contained in a document and can be naturally adapted for phenotyping from bag-of-words clinical features, where the topic--word distributions form candidate phenotypes. While LLDA assumes that the topic loadings of a document lie on the probability simplex $\Delta^{K-1}$, $\lambda$--CNMF allows each patient--condition $w_{kj}$ loading to lie in $[0,1]$. In interpreting the patient loading as a disease risk, the latter allows patients to have varying levels of disease prevalence. Also, LLDA can induce sparsity only indirectly via a hyperparameter $\beta$ of the informative prior on the topic--word distributions. While this does not guarantee sparse estimates, we obtain reasonable sparsity on LLDA estimates. We use the Gibbs sampling code from MALLET \citep{mallet} for inference. For a fair comparison to CNMF which uses an extra bias factor, we allow LLDA to model an extra topic shared by all documents in the corpus.
%The Gibbs sampling code available in MALLET \cite{mallet} is used in our experiments for LLDA inference on our data.%\footnote{\url{http://mallet.cs.umass.edu/}}.
\item \textbf{NMF with support constraints (NMF+support)}: This NMF model incorporates non--negativity and support constraints from weak supervision but not the sparsity inducing constraints on the phenotype matrix. This allows to study the effect of sparsity inducing constraints for interpretability. On the other hand, imposing sparsity without our \ground ing technique does not yield identifiable topics and hence is not studied as a baseline. %, i.e., using Algorithm \ref{alg:alg1} without constraints on $a^{(k)}$. 
 %whether weak supervision is sufficient to ensure a one-to-one mapping of conditions to the latent topics and yield robust and reliable results. %The performance of NMF without any support constraints or \grounding~was significantly worse than this baseline and are hence excluded from the discussion.
\item \textbf{Multi-label Classification (\sip)}: This baseline treats weak supervision (from ECI) as accurate labels in a fully supervised model. A sparsity inducing $\ell_1$ regularized logistic regression classifier is learned for each condition independently. The learned weight vector for each condition $k$ determines importance of clinical terms towards discriminating patients with condition $k$ and are treated as candidate phenotypes for condition $k$. %An $\ell_1$ regularized logistic regression is used as the independent classifier~\citep{park2007l1} to induce sparsity in the weights. %The regularization parameter, $\eta$, is tuned to the required sparsity of the phenotypes. 
%\item \textbf{Constrained NMF ($\mathbf{\lambda}$--CNMF)} The proposed model is trained with different initializations of $\bA$ as well as the sparsity inducing parameter $\lambda$. Each of these models is identified with the corresponding $\lambda$ parameter and is called $\lambda$--CNMF. The best performing model is chosen based on the desired level of sparsity of the learned phenotypes.
\end{asparaenum}

The weak supervision does not account for the primary diagnosis for admission in the ICU population as the ECI ignores primary diagnoses at admission~\citep{elixhauser1998comorbidity}. However, the learning algorithm can be easily modified to account for the primary diagnoses, if required by using a modified form of supervision or absorbing the effects in an additional additive term appended to the model. Nevertheless, the proposed model generates highly interpretable phenotypes for comorbidities. Finally, to mitigate the effect of local minima, whenever applicable, for each model, the corresponding algorithm was run with $5$ random initializations and results providing the lowest divergence were chosen for comparison.% and can be combined with a sparse set of raw bag-of-words features to predict $30$ day mortality. %The additional features required could be interpreted as factors explaining factors related to primary diagnoses that are predictive of patient mortality.

\subsection{Interpretability--accuracy trade--off}\label{sec:exp-sparsity}
Sparsity of the latent factors is used as a proxy for interpretability of phenotypes. Sparsity is measured as the median of the number of non--zero entries in columns of the phenotype matrix $\bA$ (lower is better). The $\lambda$ parameter in $\lambda$--CNMF controls the sparsity by imposing scaled simplex constraints on $\bA$. CNMF was trained on multiple $\lambda$ in the range of  $0.1$ to $1$.  Stronger sparsity-inducing constraints results in worse fit to the cost function. This trade--off is indeed observed in all models (see \ref{sec:spars_app} for details). % for $\lambda$--CNMF, LLDA and \sip~, respectively). %Note that for LLDA, as the sparsity can only be induced indirectly via the hyperparameter for word--topic distribution prior, the sparsity is observed to saturate at a median value $\sim400$.  
%It can be seen from the plot that the sparsity of the phenotypes can be smoothly controlled to the desired level. 
%Besides interpretability, in Section~\ref{sec:exp-phenotyping} we also observe  that the clinical relevance of phenotype candidate suffers drastically in the absence of sparsity in the phenotype. %This is intuitive as a dense phenotype (i.e. a dense estimate of $\bA$) is not discriminative of any particular condition. 
 %As seen from the figure, as more sparsity is enforced (smaller $\lambda$), accuracy deteriorates significantly.
  %We provide recommendations on the best choice of $\lambda$ in Section~\ref{sec:exp-robustness}. The sparsity of LLDA is controlled by tuning the hyperparameter of the word-topic multinomial parameters and for \sip ~via $\ell_1$ regularization. The corresponding phenotype sparsity obtained for various baselines is shown in Appendix A (see Figure~\ref{fig:sp_llda} for LLDA and Figure~\ref{fig:sp_sip} for the supervised baseline \sip).
For all models, we pick estimates with lowest median sparsity while ensuring that the phenotype candidate for every condition is represented by at least $5$ non-zero clinical terms.
%{\color{green} Include phenotype prediction performance here and pick hyper parameters based on the performance on phenotype prediction. }
%{\color{red}\textbf{In the remainder  of the paper, across all the models, we pick the parameter that achieves lowest median sparsity while ensuring that the phenotype candidate for every condition is represented by at least $5$ non-zero clinical terms. } } 
%{\color{green} Include phenotype prediction performance here and pick hyper parameters based on the performance on phenotype prediction. }

%\\
%\begin{figure}[htbp]
%  \centering 
%  \includegraphics[width=2.5in]{figures/barplot_sparsity_ad_nmf_0.pdf} 
%  \caption{Sparsity--Accuracy Trade--off}
%  \label{fig:sp_baselines} 
%\end{figure}
%Figure with 2 subplots a) sparsity vs s, b) loss values vs s
%Note about sparsity of baselines. Bar plot with sparsity.
\subsection{Clinical relevance of phenotypes}\label{sec:exp-phenotyping}
\begin{figure}[htb]
\small 
\centering
\includegraphics[width=0.65\textwidth]{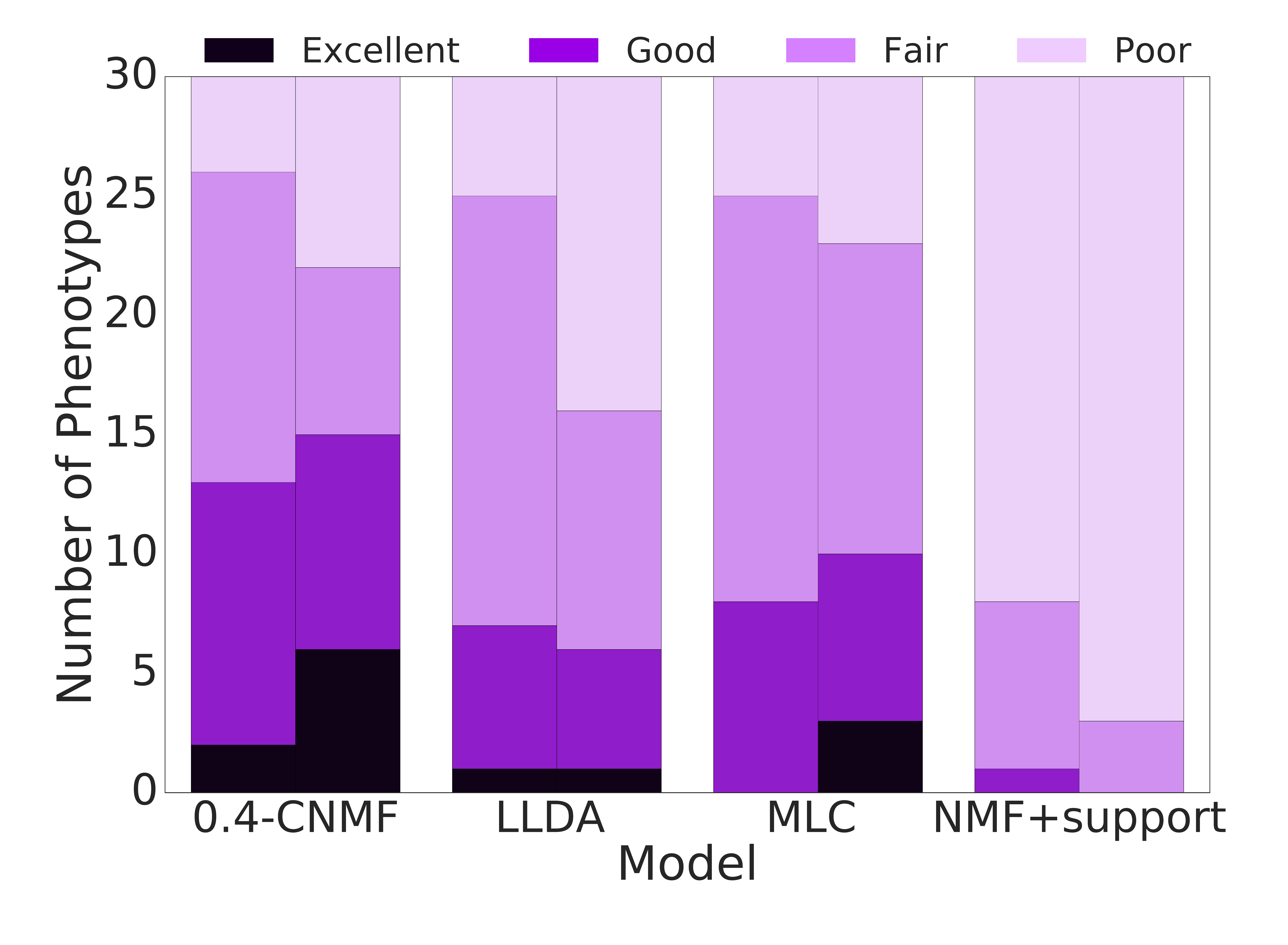}
\vspace{-4mm}
\caption{Qualitative Ratings from Annotation: The two bars represent the  ratings  provided by the two annotators. Each bar is a  histogram of the scores for the $30$ comorbidities sorted by scores. }
\label{fig:ratings_bargraph}
\end{figure}
%bar plot with histogram of scores for each phenotype and each baseline
%As suggested before, clinical interpretability is of critical significance for phenotyping algorithms. 
%To evaluate clinical relevance of the phenotypes, 

We requested two clinicians to evaluate the candidate phenotypes based on the top $15$ terms learned by each model. The ratings were requested on a scale of $1$ (poor) to $4$ (excellent). The experts were asked to rate based on whether \textit{the terms are relevant towards the corresponding condition and whether the terms are jointly discriminative of the  condition}. 
%\begin{minipage}{0.7\textwidth}
%  \includegraphics[width=0.7\textwidth]{figures/phenotype_ratings.pdf}
%\captionof{figure}{Qualitative Ratings from Annotation: The figure shows the distribution of ratings provided by both annotators (corresponding to each column) for all models.}
%\label{fig:ratings_bargraph}
%\end{minipage}
%\begin{minipage}{0.3\textwidth}
%  \begin{tabular}{|l|l|l|l|l|}
%\hline
% &  \bf 0.4--CNMF &  LLDA  &   \sip  &  NMF \\ \hline
%\bf 0.4--CNMF &  \bf 0 & \bf 28  & \bf 20 &  \bf 44 \\ \hline
% LLDA  & \bf 7 & 0  & 12 & 35 \\ \hline
% \sip   & \bf 6 & 21 &0  & 42 \\ \hline
% NMF+support & \bf 1 &0  & 1 & 0 \\ \hline
%\end{tabular}
%\captionof{table}{Relative Rankings Matrix: Each row of the table is the number of times the model along the row was rated \textit{strictly} higher than the model along the corresponding column by clinical experts, e.g., column $3$ in row $2$ implies that $12$ times LLDA was rated better than \sip~(across all ratings of $2$ experts for $30$ conditions).}
%\label{tab:ratings}
%\end{minipage}
Figure~\ref{fig:ratings_bargraph} shows the summary of qualitative ratings obtained for all models. For each model, we show two columns (corresponding to two experts). The stacked bars show the histogram of the ratings for the models. Nearly $50\%$ of the phenotypes learned from our model were rated `good' or better by both annotators. In contrast, NMF with support constraints but \emph{without} sparsity inducing constraints hardly learns clinically relevant phenotypes.
\begin{table}[htb]{
\small
\centering
\begin{tabular}{|l|l|l|l|l|}
\hline
 &  \bf 0.4--CNMF &  LLDA  &   \sip  &  NMF \\ \hline
\bf 0.4--CNMF &  \bf 0 & \bf 28  & \bf 20 &  \bf 44 \\ \hline
 LLDA  & \bf 7 & 0  & 12 & 35 \\ \hline
 \sip   & \bf 6 & 21 &0  & 42 \\ \hline
 NMF+support & \bf 1 &0  & 1 & 0 \\ \hline
\end{tabular}
\caption{Relative Rankings Matrix: Each row of the table is the number of times the model along the row was rated \textit{strictly} better than the model along the column by clinical experts, e.g., column $3$ in row $2$ implies that LLDA was rated better than \sip~$12$ times over all conditions by all experts.}
\label{tab:ratings}}
\end{table}
The proposed model $0.4$--CNMF also received significantly higher number of `excellent' and `good' ratings from both experts. Although LLDA and \sip~estimate sparse phenotypes, they are not at par with $\lambda$--CNMF. %We also study the number of times a given model is rated strictly better than other models. 
Table~\ref{tab:ratings} shows a summary of relative rankings for all models. Each cell entry shows the number of times the model along the corresponding row was rated \textit{strictly better} than that along the column. $0.4$--CNMF is better than all three baselines.  The supervised baseline MLC outperforms LLDA even though LLDA learns comorbidities jointly suggesting that the simplex constraint imposed by LLDA may be restrictive.

\begin{figure}[tb]
\small
  \centering
  \includegraphics[width=0.8\textwidth]{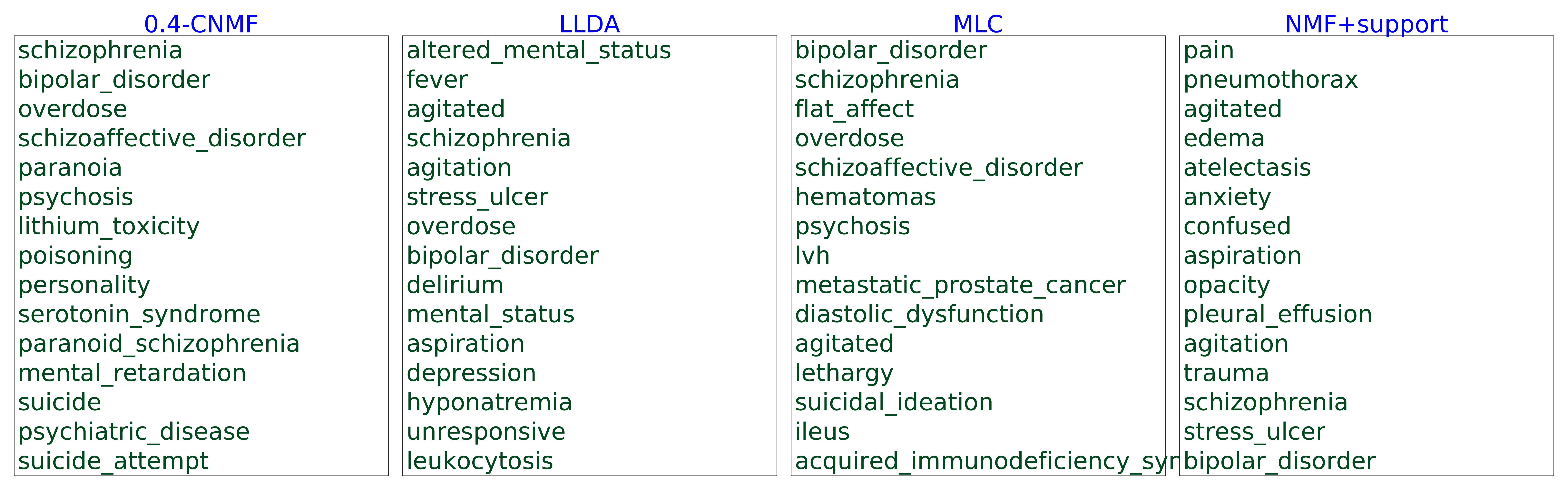}
  \caption{Phenotypes learned for `Psychoses' (words are listed in order of importance)}
  \label{fig:psychoses}
\end{figure}

\begin{figure}[tb]
\small
  \centering
  \includegraphics[width=0.8\textwidth]{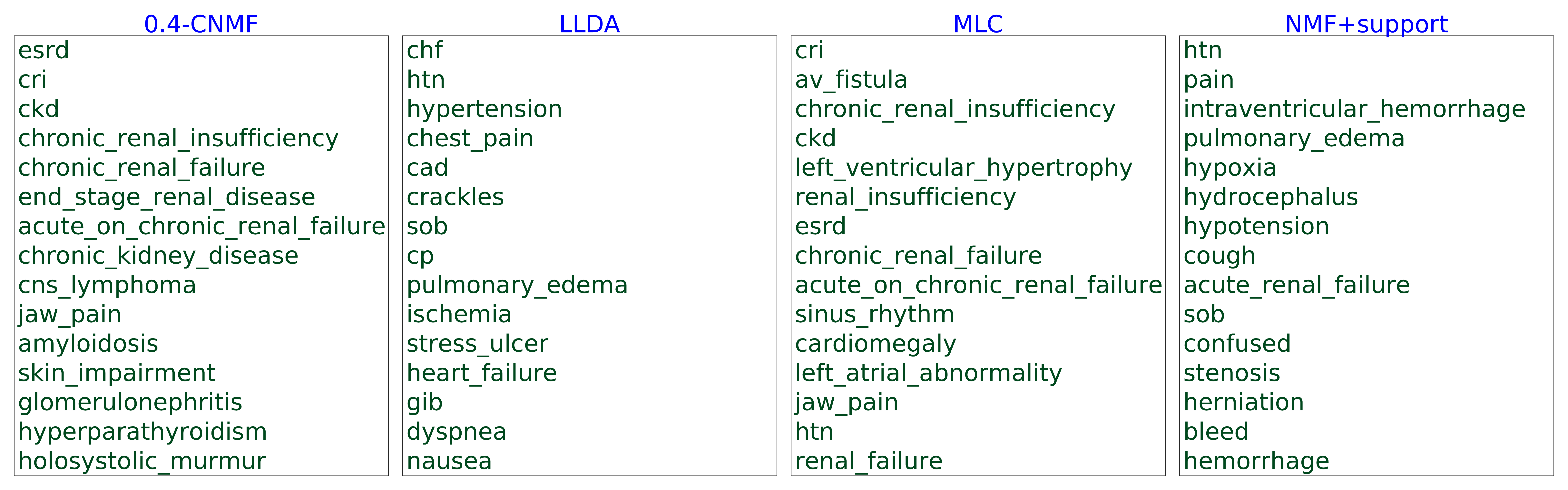}
  \caption{Phenotypes learned for `Hypertension'}
  \label{fig:hypertension}
\end{figure}

Figure~\ref{fig:psychoses} %and~\ref{fig:liver_disease} 
is an example of a phenotype (top 15 terms) learned by all models for psychoses. For this condition, the proposed model was rated ``excellent" and strictly better than both LLDA and MLC by both annotators while LLDA and MLC ratings were tied. However, the phenotype for Hypertension (in Figure~\ref{fig:hypertension}) learned by $0.4$--CNMF has more terms related to `Renal Failure' or `End Stage Renal Disease' rather than hypertension. One of our annotators pointed out that  ``Candidate 1 is a fairly good description of renal disease, which is an end organ complication of hypertension'', where the anonymized Candidate 1 refers to $0.4$--CNMF. Exploratory analysis suggests that hypertension and renal failure are the most commonly co-occurring set of conditions. Over 93\% of patients that have hypertension (according to ECI) also suffer from Renal Failure. Thus, our model is unable to distinguish between highly co-occurring conditions. Other baselines were also rated poorly for hypertension, while LLDA was rated only slightly better. More examples of phenotypes are provided in \ref{sec:samples}.
%Figure~\ref{fig:cnmf_phenotypes_best} also shows the top 15 terms learned by $0.4$--CNMF for Liver Disease and Psychoses. The proposed model was rated highly in terms for both these conditions by both annotators. Sample phenotypes for the baseline models have been included in Appendix B.
%\begin{figure}[htbp]
%\small
%  \centering 
%  \begin{subfigure}{.66\textwidth}
%  \includegraphics[width=\textwidth]{figures/cnmf_best_ratings.pdf}
%  \caption{Clinically meaningful sample phenotypes}
%  \label{fig:cnmf_phenotypes_best} 
%  \end{subfigure}%
%  \begin{subfigure}{.33\textwidth}
%  \includegraphics[width=\textwidth]{figures/cnmf_failure_case.pdf} 
% \caption{Non-discriminative phenotype}
%  \label{fig:cnmf_phenotypes_worst} 
%  \end{subfigure}
%  \label{fig:cnmf_phenotypes}
%  \caption{Sample phenotypes $0.4$--CNMF}
%\end{figure}
%Figure~\ref{fig:cnmf_phenotypes} shows sample phenotypes obtained for the proposed model for Liver Disease, Psychoses and Hypertension.
%We observe that our model outperforms LLDA in clinical relevance of phenotypes in spite of obtaining sparse estimates of phenotypes. 
%\input{temp}
\subsection{Mortality prediction}\label{sec:exp-pred}

To quantitatively evaluate the utility of the learned phenotypes, we consider the $30$ day mortality prediction task. We divide the EHR into $5$ cross-validation folds of $80\%$ training and $20\%$ test patients. As this is an imbalanced class problem, the training--test splits are stratified by mortality labels.
For each split, all models were applied on the training data to obtain phenotype candidates $\tilde{\bA}$ and feature biases $\tilde{b}$. For each model, the patient loadings $\tilde{\bW}$ along the respective phenotype space $(\tilde{\bA},\tilde{b})$ are used as features to train a logistic regression classifier for mortality prediction. %The patient loadings for train  $\bW_{\text{train}}$ and test $\bW_{\text{test}}$ patients are obtained by projections of $\bX_{\text{train}}$ and $\bX_{\text{test}}$, respectively, onto the latent space of phenotypes  $(\tilde{\bA},\tilde{b})$ learned on training data.  
 For CNMF and NMF+support, these are obtained as  $\bW_\text{train/test}={\text{argmin}}_{\small \bW\in[0,1]^{K\times N}} \mathcal{D}( \tilde{\bA}\bW+\tilde{b}\mathbbm{1}^\top,\bX_\text{train/test})$ for fixed $(\tilde{\bA},\tilde{b})$. For LLDA, these are obtained using Gibbs sampling with fixed topic--word distributions. For MLC, the predicted class probabilities of the comorbidities are used as features. Additionally, we train a logistic regression classifier using the full EHR matrix as features. 
 
We clarify the following points on the methodology:
\begin{inparaenum}[(1)]
\item $\tilde{\bA}$ is learned on the patients in the training dataset only, hence there is no information leak from test patients into  training. 
 \item Test patients' comorbidities from ECI are \textit{not} used as support constraints on their loadings. 
 \item Regularized logistic regression classifiers are used to learn models for  mortality prediction. The regularization parameters are chosen via grid-search.
\end{inparaenum}
%More precisely, let $\tilde{\bA}$ be the phenotype matrix estimated from the training population, $\tilde{b}$, the estimated bias, and let $\bW_{train}$ be the patient loadings matrix from $\bX_{train}$, the EHR consisting of 80\% of the population. Let $\bX_{test}\in\mathbb{R}_+^{d\times N_{test}}$ be the EHR for the unseen patient population whose 30-day mortality is to be predicted. We estimate the new patients' loadings as follows
%\begin{equation}\label{eq:nmf}
%\begin{aligned}
%\bW_{test} = &\underset{\bW\ge 0}{\text{argmin}}&& \mathcal{D}( \tilde{\bA}\bW+\tilde{b}\mathbbm{1}^\top,\bX_{test}) & \\ %+\lambda \|\bW\|_F^2& \\
%&&& \bW \in [0,1]^{K \times N_{test}},
%\end{aligned}
%\end{equation}
%$\bW_{train}$ estimated from Algorithm~\ref{alg:alg1} are the unseen test patient population, patient loadings are estimated by projecting the respective clinical notes to the phenotype space. 
%Note that no information about patients' comorbidities is used while estimating the loadings on an unseen population. Thus the estimated phenotypes automatically provide an estimate of risk for all chronic conditions for an unseen patient population. 
\begin{table}[t]
\centering
\small
\begin{tabular}{|l|l|l|l|l|}
\hline
& Model       & AUROC           & Sensitivity    & Specificity    \\ \hline
1. &{$0.4$--CNMF}          &  0.63(0.02)  & 0.59(0.04) & 0.62(0.03)  \\
2.  & NMF+support          & 0.52(0.02)  & 0.56(0.13) & 0.51(0.14)  \\
3.  & LLDA &  0.64(0.02) & 0.62(0.03) & 0.61(0.05)  \\
4.  & {\sip} &  0.66(0.01) & 0.62(0.06) & 0.62(0.05) \\
\bf 5.  & \bf Full EHR & \bf 0.72(0.02) & \bf 0.69(0.02) & \bf 0.63(0.04) \\ \hline
\bf 6.  & \bf CNMF+Full EHR ($\ell_1$, $C=0.1$)& \bf 0.72(0.02) & \bf 0.61(0.09) & \bf 0.71(0.07) \\\hline
\end{tabular}
\caption{$30$ day mortality prediction: $5$--fold cross-validation performance of logistic regression classifiers. Classifiers for $0.4$--CNMF and competing baselines (NMF+support, LLDA, \sip) were trained on the $30$ dimensional phenotype loadings as features. Full EHR denotes the baseline classifier ($\ell_1$-regularized logistic regression) using full $\sim3500$ dimensional EHR as features. CNMF+Full EHR denotes the  performance of the $\ell_1$-regularized classifier learned on Full EHR augmented with CNMF features (hyperparameter was manually tuned to match performance of the Full EHR model). 
\label{tab:mortality_pred}}
\end{table}

The performance of the above baselines trained on $\ell_2$ regularized logistic regression over a $5$-fold cross-validation is reported in  Table~\ref{tab:mortality_pred}: rows $1$--$5$. The classifier trained on the full EHR unsurprisingly outperforms all baselines as it uses richer high dimensional information. %However, as noted earlier full EHR data is not easily interpretable or suitable for large scale clinical trials. 
All phenotyping baselines, except NMF+support, show comparable performance on mortality prediction which in spite of learning on a small number of $30$ features, is only slightly worse than predictive performance of full EHR with $\sim 3500$ features.
\paragraph{Augmented features for mortality prediction (CNMF+Full EHR)} %As noted previously, the weak supervision used in our method ignores the primary cause of admission for patients. 
Unsurprisingly, Table~\ref{tab:mortality_pred} suggests that the high dimensional EHR data has additional information towards mortality prediction which are lacking in the $30$ dimensional features generated via phenotyping. To evaluate whether this additional information can be captured by CNMF if augmented with a small number of raw EHR features, we train a mortality prediction classifier using $\ell_1$ regularized logistic regression on CNMF features/loadings combined with raw {bag--of--words} features, with parameters tuned to match the performance of the full EHR model. The results are reported in the final row of  Table~\ref{tab:mortality_pred}. 

\begin{figure}[htb]
\centering
\includegraphics[width=0.35\textwidth,angle=0]{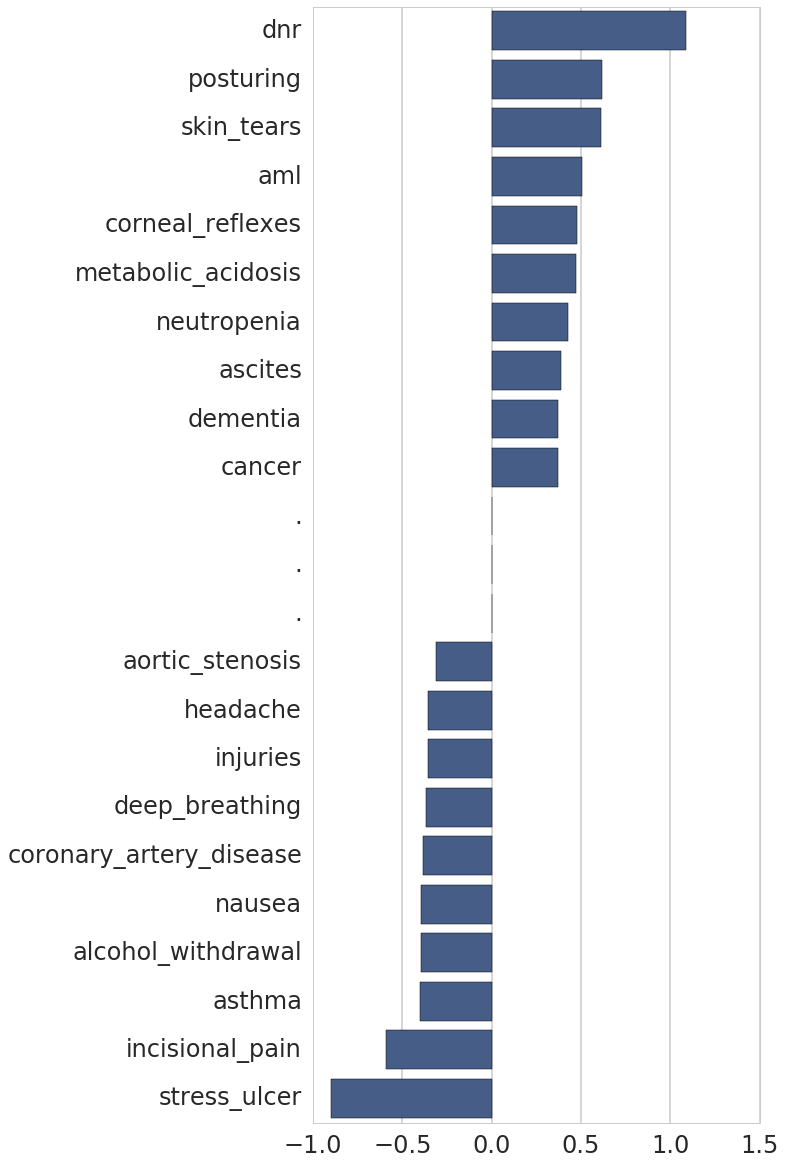}
~
\includegraphics[width=0.33\textwidth,angle=0]{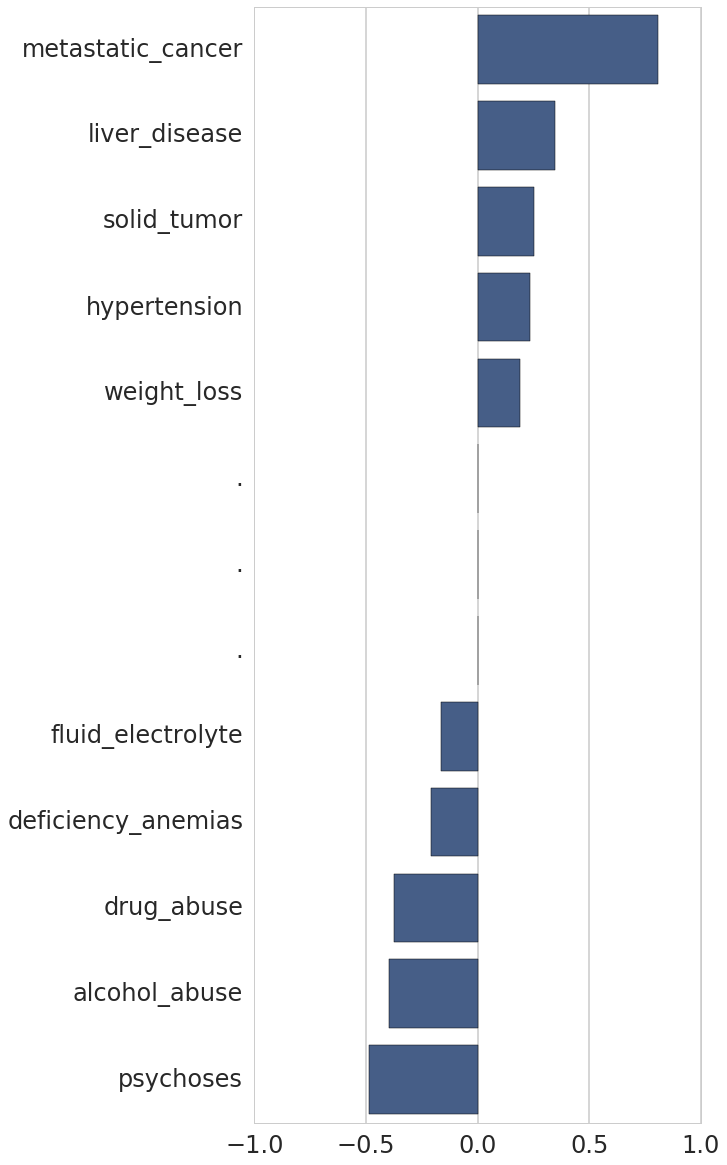}
\caption{Top magnitude weights  on (a) EHR and (b) CNMF features in CNMF+Full EHR classifier}
\label{fig:aug_cnmf_feat2}
\end{figure}

In exploring the weights learned by the classifier for all features, we observe that only $8.3\%$ of the features corresponding to raw EHR based \emph{bag-of-words} features have non--zero weights. %Further, the average magnitude of the non--zero weights corresponding to EHR features is 0.13 and are comparable to those of CNMF features (0.17). 
This suggests that comorbidities capture significant amount of predictive information on mortality and achieve comparable performance to full EHR model with a small number of additional terms. See Figure~\ref{fig:aug_cnmf_feat1} in Appendix showing the weights learned by the classifier for all features. 
Figure~\ref{fig:aug_cnmf_feat2} shows comorbidities and EHR terms with top magnitude weights learned by the CNMF+full EHR classifier. For example, it is interesting to note that the top weighted EHR term -- dnr or `Do Not Resuscitate' is not indicative of any comorbidity but is predicitive of patient mortality.

%\clearpage

\section{Discussion and Related Work}\label{sec:discussion}
%{\color{blue}Yuan Luo’s 2015 JAMIA paper.}
%{\color{blue} ``Weakness 1: Clinical interpretability is nice, but not forsaking utility. The authors really only evaluate on one downstream task: 30-day mortality prediction. Yet on that task the model doesn’t really perform that well. I’m a bit surprised the authors didn’t evaluate on phenotype classification using held-out data, as that seems like the obvious task."

%``the explanation of the results is useful although somewhat overdetailed. I'm afraid that there boxes of multiple phenotypes as typified by figure 2 are poorly analyzed ultimately and not overly convincing. certainly it would be useful to be able to phenotype patient's from the medical records however I do not feel that this approaches convincing mildly better than a medical student reading the notes. If this is the root the end result of a prolonged process. Arthur's do not hold out hope of improvement. I feel this paper is perhaps a little over concluded based on the weakness of their results. It is certainly an interesting approach although I feel it is somewhat naïve from a clinical perspective--{\color{black} Note that the results are on one set of features only but the methodology can be applied on datasets with richer features}

%``Lack of qualitative comparisons, especially how the discovered phenotypes can be turned into actionable insights" 

%}

%Machine learning methods for automated phenotyping can be classified based on the type of data used, supervision, algorithms etc. 
Supervised learning methods like \citet{carroll2011naive,kawaler2012learning,Chen:2013es} or deep learning methods~\citep{lipton2015learning,kale2015causal, henaoelectronic} for EHR driven phenotyping require expert supervision. Although unsupervised methods like NMF~\citep{anderson2014non} and non--negative tensor factorization \citep{kolda2009tensor,harshman1970foundations} are inexpensive alternatives~\citep{Ho2014:JBI, Ho2014:BIH, Ho2014:KDD,Luo1009},  %The NMF formulation proposed by~\citet{anderson2014non} focus on a single disease and the learned factors have to be \emph{post-hoc} evaluated for identifiability. Further, 
%Most methods that build on NMF primarily use the least squares loss functions (implicitly assuming continuous valued data with Gaussian--like noise) in the optimization procedure and do not extend well to count data, a popular representation of text data, daignosis codes and medications. 
%In comparison the model proposed in this paper is more general as the choice of divergence can be made appropriately according to the data types of interest. 
%A recent line of work by \citet{Ho2014:JBI, Ho2014:BIH, Ho2014:KDD} has illustrated the promise of tensor factorization methods \citep{kolda2009tensor,harshman1970foundations} to generate phenotypes with minimal human supervision. 
%However, %most EHR data are currently available in flat file formats and 
%learning from higher order tensor models requires exponential storage and computational costs, limiting their scalability. 
% In another unsupervised method,~\citet{henaoelectronic} use Deep Poisson Factorization models,% a framework that learns topic models~\citep{blei2003latent} using poisson factorization modules. This work 
%to jointly models multiple modalities of the EHR in a probabilistic graphical model framework. %The learned latent factors are trained to be discriminative of multiple chronic conditions. Unfortunately, few of these methods focus on clinical interpretability. 
they pose challenges with respect to identifiability, interpretability and computation, limiting their scalability. %For example, although PCA provides a unique solution to the underlying formulation, the factors learned need not be clinically interpretable. 
%To counter these issues, 
%Reducing manual intervention while achieving identifiability for phenotyping has received some success recently. 

%Identifiable phenotyping with minimal expert intervention has received some success.
Most closely related to our paper is work by \citet{halpern2016mlhc} which is a semi-supervised algorithm for learning the joint distribution on conditions, requiring only that a domain expert specify one or more `anchor' features for each condition (no other labeled data). An `anchor' for a condition is a set of clinical features that when present are highly indicative of the target condition, but whose absence is not a strong label for absence of the target condition \citep{halpern2014using, halpern2016electronic}. For example, the presence of insulin medication is highly indicative of diabetes, but the converse is not true. \citet{joshi2015simultaneous} use a similar supervision approach for comorbidities prediction. Whereas the conditions in \citet{halpern2016mlhc} are binary valued, in our work they are real-valued between 0 and 1. Furthermore, we assume that the {\em support} of the conditions is known in the training data. 

Our approach achieves identifiability using support constraints to \ground~the latent factors and interpretability using sparsity constraints.
The phenotypes learned are clinically interpretable and predictive of mortality when augmented with a sparse set of raw \emph{bag-of-words} features on unseen patient population. %The proposed model is compared with strong baselines in terms of the sparsity and clinical relevance from experts. 
The model outperforms baselines in terms of clinical relevance according to experts and significantly better than the model which includes supervision but no sparsity constraints. The proposed method can be easily extended to other non--negative data to obtain more comprehensive phenotypes. However, it was observed that the algorithm does not discriminate between frequently co--occurring conditions, e.g. renal failure and hypertension. Further, the weak supervision (using ECI) does not account for the primary diagnoses of admission. Additional model flexibility to account for a primary condition in explaining the observations could potentially improve performance. %Alternately, a different form of supervision that accounts for primary diagnoses of patients from may be used. 
%Finally, with 48 hours of clinical data, our model is not entirely robust to initialization. Preliminary experiments (omitted here for brevity) suggest that with more data, the \ground ing constraints and sparsity are strong enough to decrease the number of local minima the algorithm can potentially converge to. 
Addressing the above limitations along with quantitative evaluation of risk for disease prediction, and understanding conditions for uniqueness of phenotyping solutions are interesting areas of follow-up work. %Preliminary experiments (omitted here for brevity) agree with this expectation. 

\clearpage
\section*{Acknowledgements}
We thank Dr. Saul Blecker and Dr. Stephanie Kreml for their qualitative evaluation of the computational phenotypes. SJ, SG and JG were supported by NSF: SCH \#1418511. DS was supported by NSF CAREER award \#1350965. We also thank Yacine Jernite for sharing a code used in preprocessing clinical notes.
%{\color{blue}TODO: Other acknowledgements}

\bibliography{phenotyping_cnmf,healthcare,bibliography}
\clearpage
\appendix
\renewcommand{\thesection}{~\Alph{section}}
\section{Phenotype Sparsity}\label{sec:spars_app}

\begin{figure}[htbp]
  \centering 
  \small
  \includegraphics[width=\textwidth]{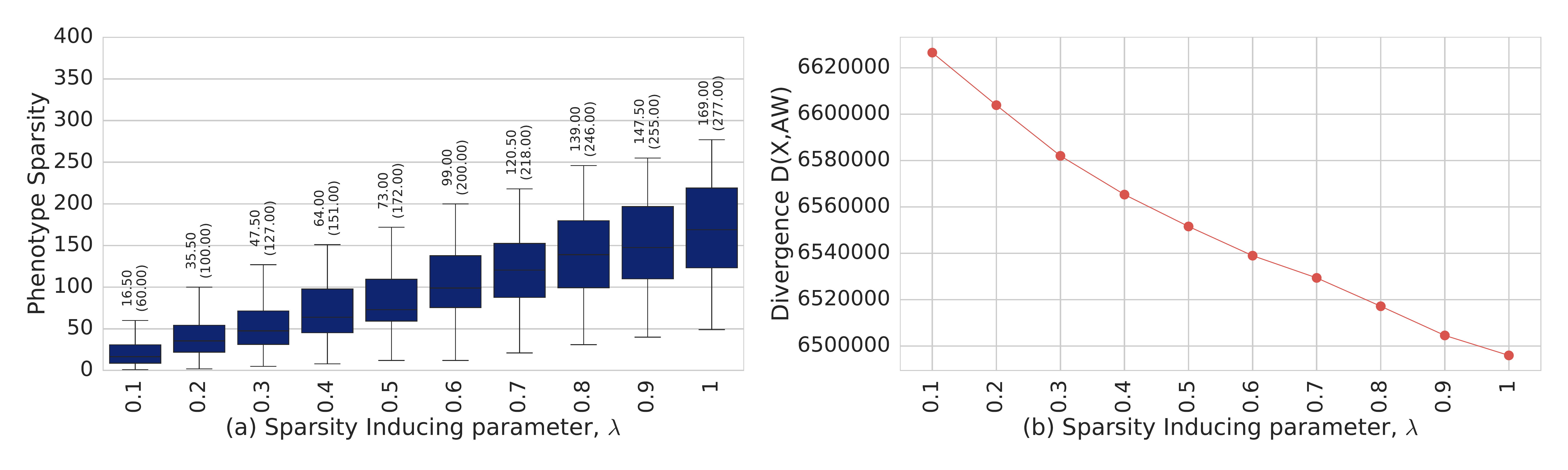} 
  \caption{Sparsity--Accuracy Trade--off. Sparsity of the model is measured as the median of the number of non-zero entries in columns of the phenotype matrix $\bA$. (a) shows a box plots of the median sparsity across the $30$ chronic conditions for varying $\lambda$ values. The median and third--quartile values are explicitly noted on the plots. (b) divergence function value of the estimate from Algorithm~\ref{alg:alg1} plotted against $\lambda$ parameter.  \label{fig:sp_acc} }
\end{figure}

\begin{figure}[b]
\centering
\begin{subfigure}{.5\textwidth}
  \centering
  \includegraphics[width=\linewidth]{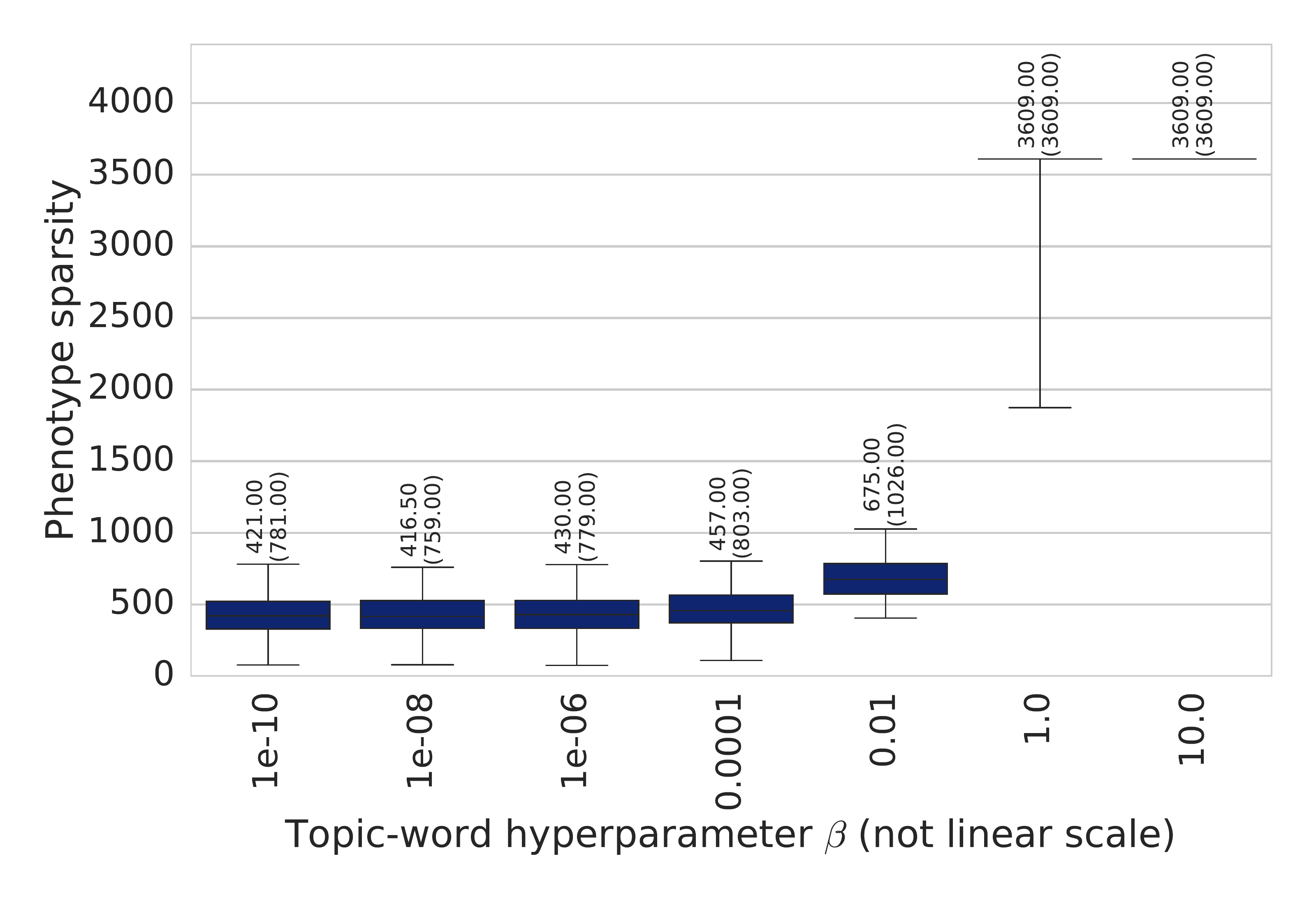}
  \caption{LLDA}
  \label{fig:sp_llda}
\end{subfigure}%
\begin{subfigure}{.5\textwidth}
  \centering
  \includegraphics[width=\linewidth]{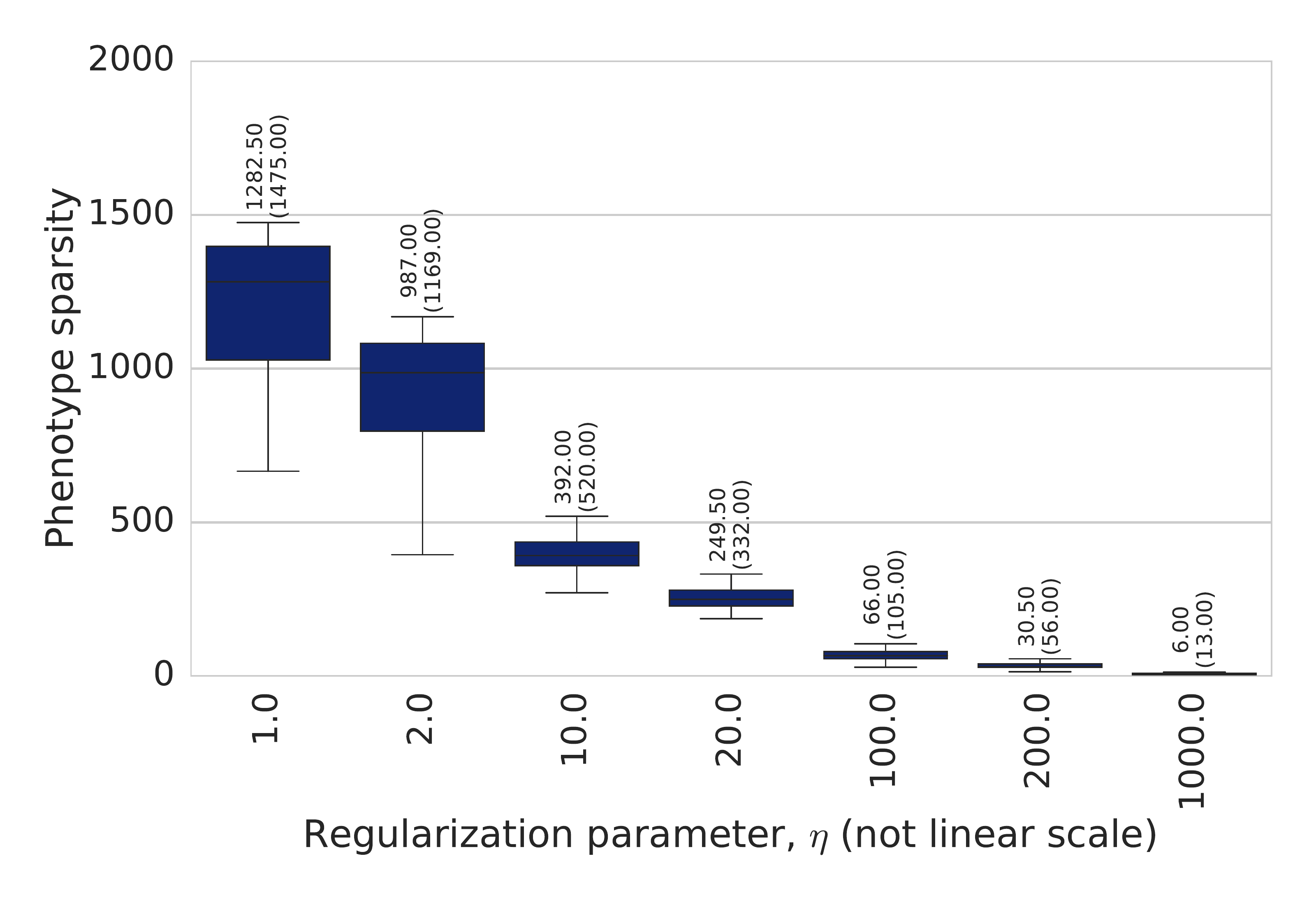}
  \caption{\sip}
  \label{fig:sp_sip}
\end{subfigure}
\caption{Phenotype sparsity for baseline models}
%\label{fig:test}
\end{figure}

As suggested in Section~\ref{sec:exp-sparsity}, there is an inherent tradeoff between fit to the cost function and desired sparsity. The trade-off is made explicit for $\lambda$--CNMF in Figure~\ref{fig:sp_acc}. The sparsity of LLDA is controlled by tuning the hyperparameter ($\beta$) of the word-topic multinomial parameters~\citep{blei2003latent} and for \sip ~via the $\ell_1$ regularization parameter $\eta$. A smaller value of $\beta$ ensures that the word-topic probabilities are sparse. As the value of $\beta$ is increased, sparsity decreases (i.e. number of non-zero elements increases). For logistic regression (used by \sip), as the $\ell_1$ regularization parameter increases, sparsity increases. Figure~\ref{fig:sp_llda} demonstrates the sparsity of the estimated phenotypes for LLDA and Figure~\ref{fig:sp_sip} shows that of logistic regression. We choose phenotypes obtained at $\beta =  1 \times 10^{-8}$ and $\eta = 100$ for qualitative annotation. The parameters were chosen to achieve the lowest median sparsity while ensuring that for each chronic condition, the corresponding phenotype candidate is represented by at least $5$ non-zero clinical terms. Our fourth baseline (NMF + support) did not estimate sparse phenotypes and does not have a tuneable sparsity parameter (but were nevertheless annotated for qualitative evaluation). The proposed model provides the best sparsity among all baselines.

%\begin{figure}[htbp]
%\centering
%\begin{subfigure}{.5\textwidth}
%  \centering
%  \includegraphics[width=\linewidth]{figures/llda_48_full_pheno_sparsity.pdf}
%  \captionof{figure}{Phenotype sparsity}
%  \label{fig:sp_llda}
%\end{minipage}%
%\begin{minipage}{.5\textwidth}
%  \centering
%  \includegraphics[width=\linewidth]{figures/logistic_48_full_sparsity.pdf}
%  \captionof{figure}{Another figure}
%  \label{fig:sp_sip}
%\end{minipage}
%\end{figure}

\section{Sample Phenotypes for Baseline Models}\label{sec:samples}
\begin{figure}[!htbp]
\small
  \centering
  \includegraphics[width=\textwidth]{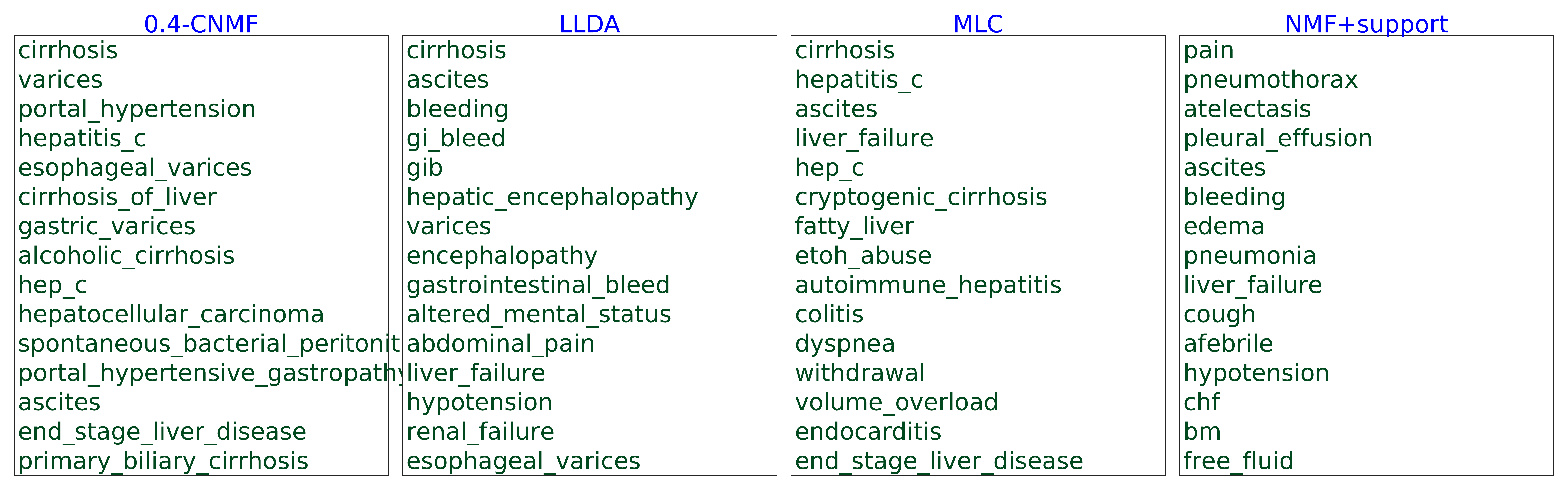}
  \caption{Learned Phenotypes for Liver Disease}
  \label{fig:liver_disease}
\end{figure}

\begin{figure}[!htbp]
\small
  \centering
  \includegraphics[width=\textwidth]{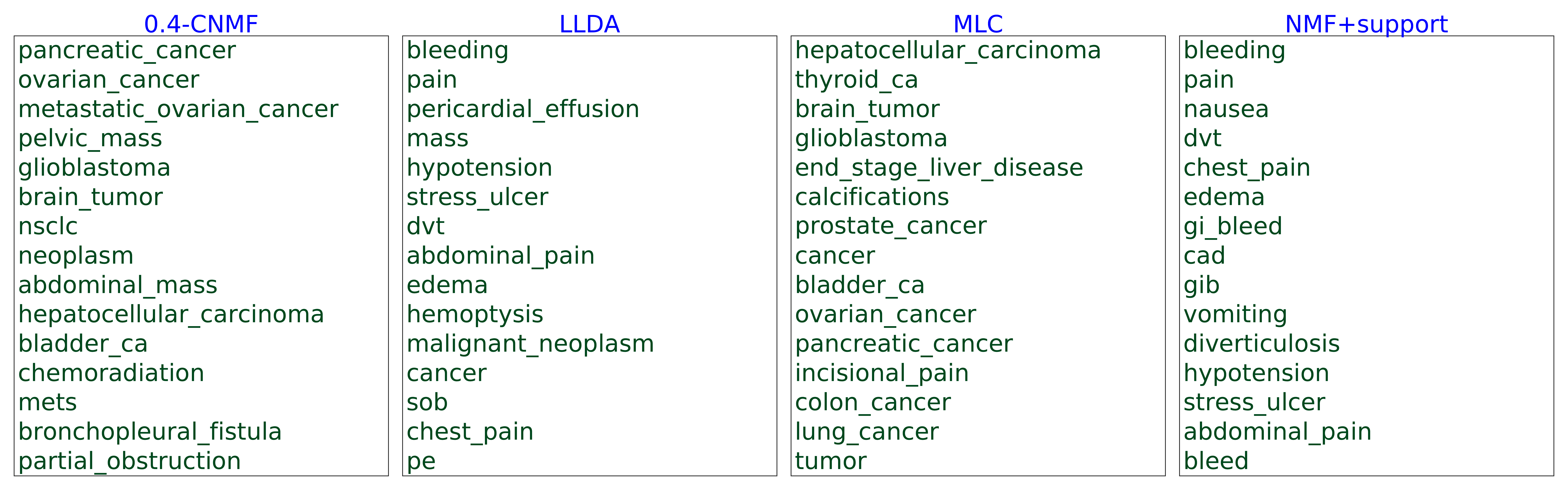}
  \caption{Learned Phenotypes for Solid Tumor}
  \label{fig:solid_tumor}
\end{figure}

\begin{figure}[!htbp]
\small
  \centering
  \includegraphics[width=\textwidth]{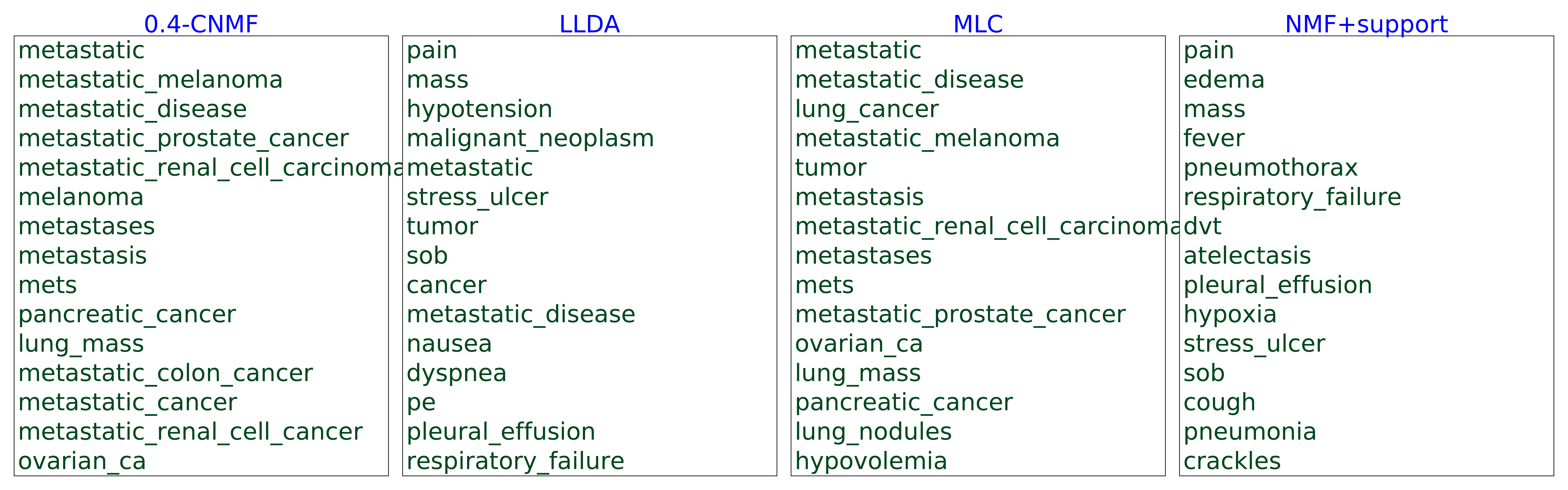}
  \caption{Learned Phenotypes for Metastatic Cancer}
  \label{fig:metastatic_cancer}
\end{figure}

\begin{figure}[!htbp]
\small
  \centering
  \includegraphics[width=\textwidth]{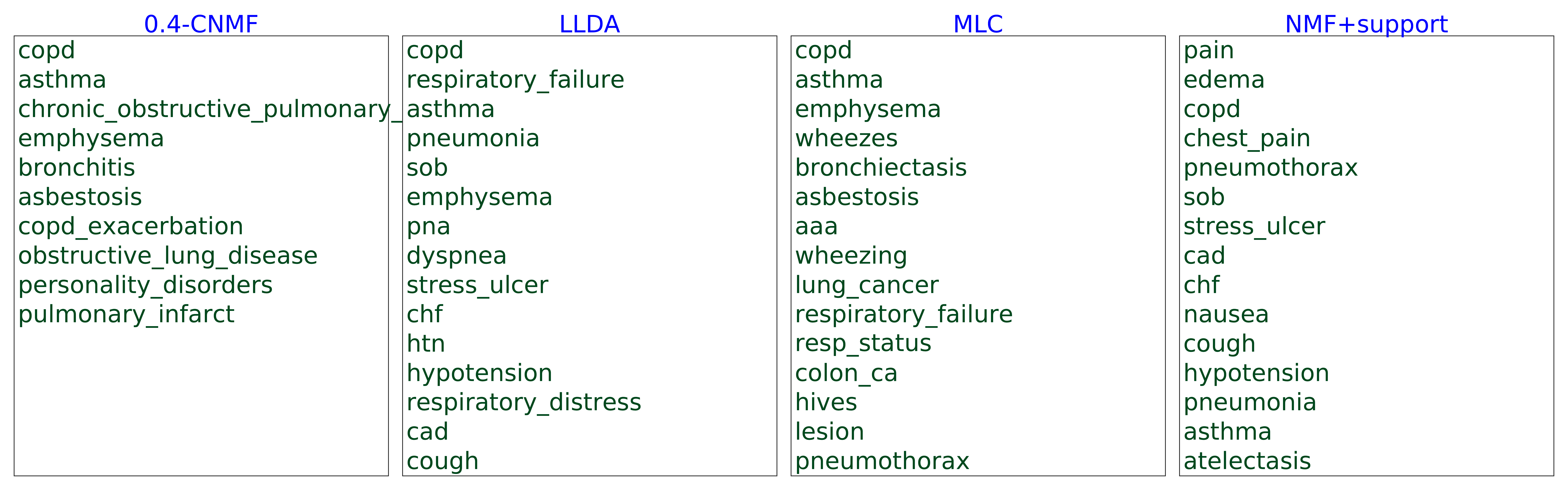}
  \caption{Learned Phenotypes for Chronic Pulmonary Disorder}
  \label{fig:chronic_pulmonary}
\end{figure}

\begin{figure}[!htbp]
\small
  \centering
  \includegraphics[width=\textwidth]{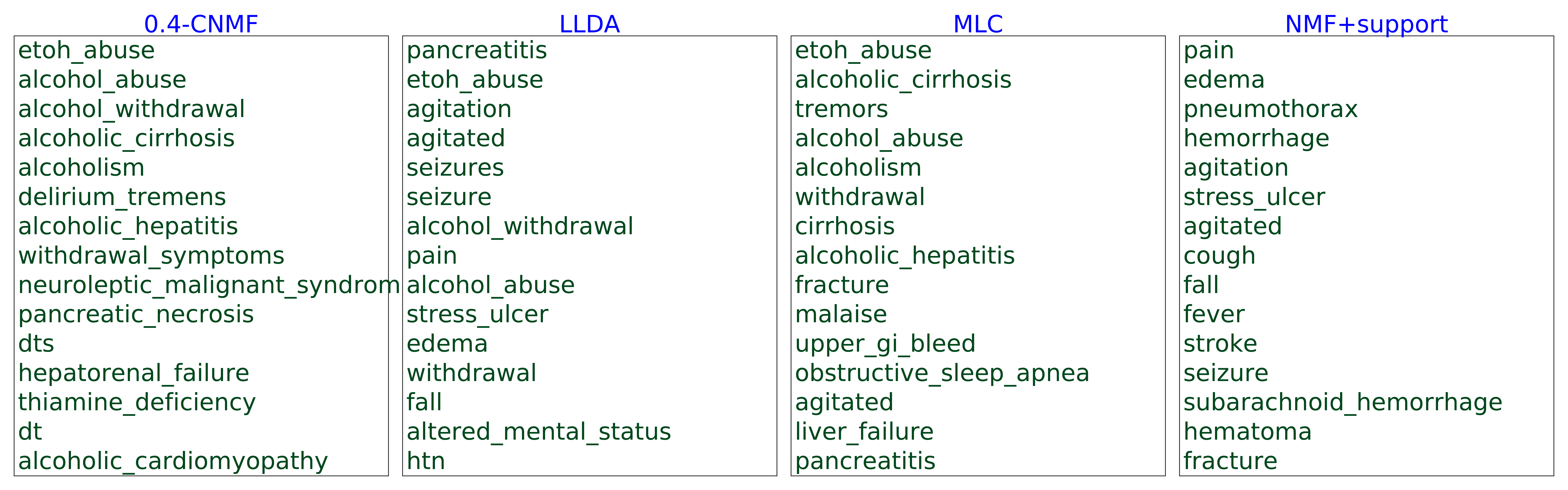}
  \caption{Learned Phenotypes for Alcohol Abuse}
  \label{fig:alcohol_abuse}
\end{figure}

\begin{figure}[!htbp]
\small
  \centering
  \includegraphics[width=\textwidth]{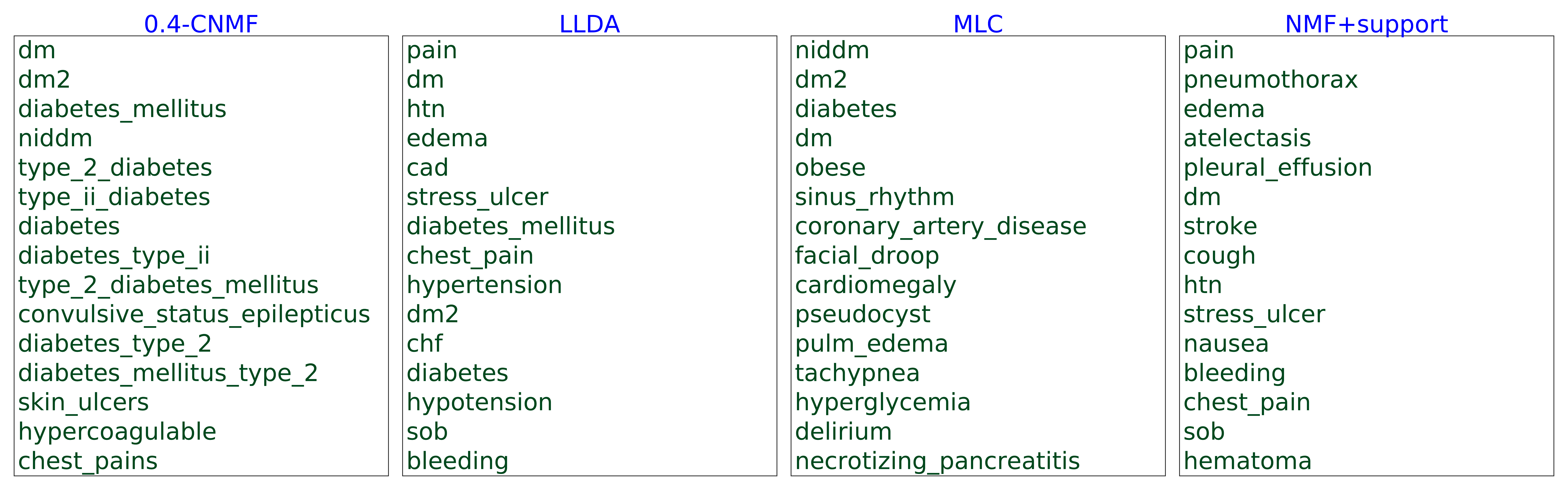}
  \caption{Learned Phenotypes for Diabetes Uncomplicated}
  \label{fig:diabetes_uncomplicated}
\end{figure}

\begin{figure}[!htbp]
\small
  \centering
  \includegraphics[width=\textwidth]{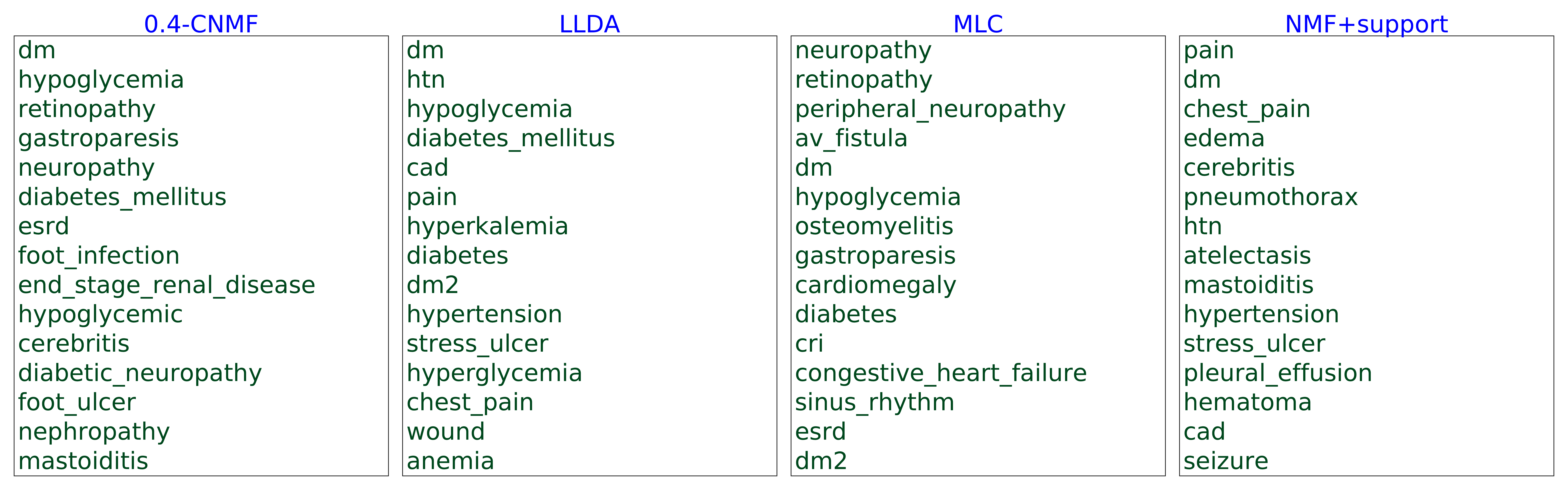}
  \caption{Learned Phenotypes for Diabetes Complicated}
  \label{fig:diabetes_complicated}
\end{figure}

\begin{figure}[!htbp]
\small
  \centering
  \includegraphics[width=\textwidth]{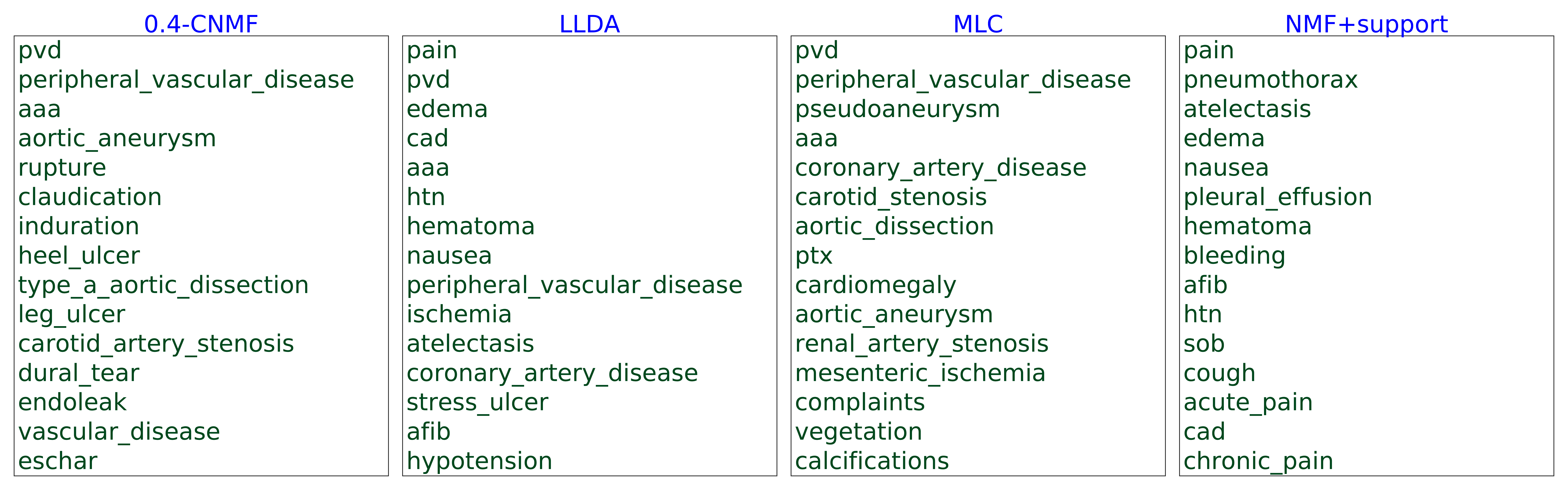}
  \caption{Learned Phenotypes for Peripheral Vascular Disorder}
  \label{fig:peripheral_vascular}
\end{figure}

\begin{figure}[!htbp]
\small
  \centering
  \includegraphics[width=\textwidth]{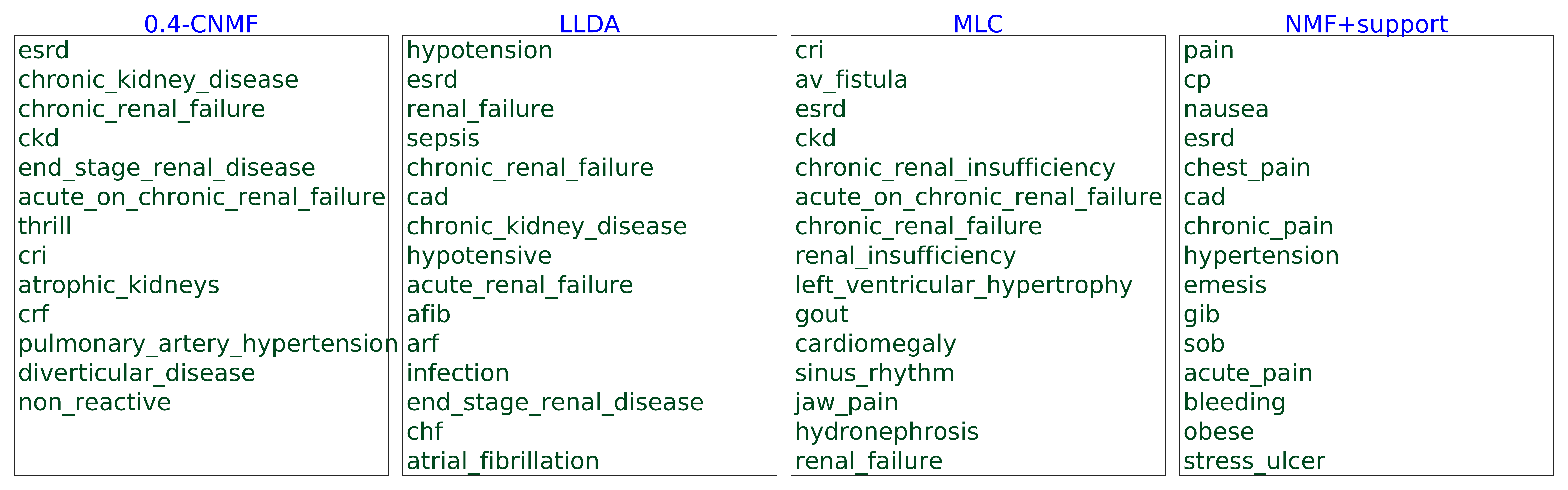}
  \caption{Learned Phenotypes for Renal Failure}
  \label{fig:renal_failure}
\end{figure}

\begin{figure}[!htbp]
\small
  \centering
  \includegraphics[width=\textwidth]{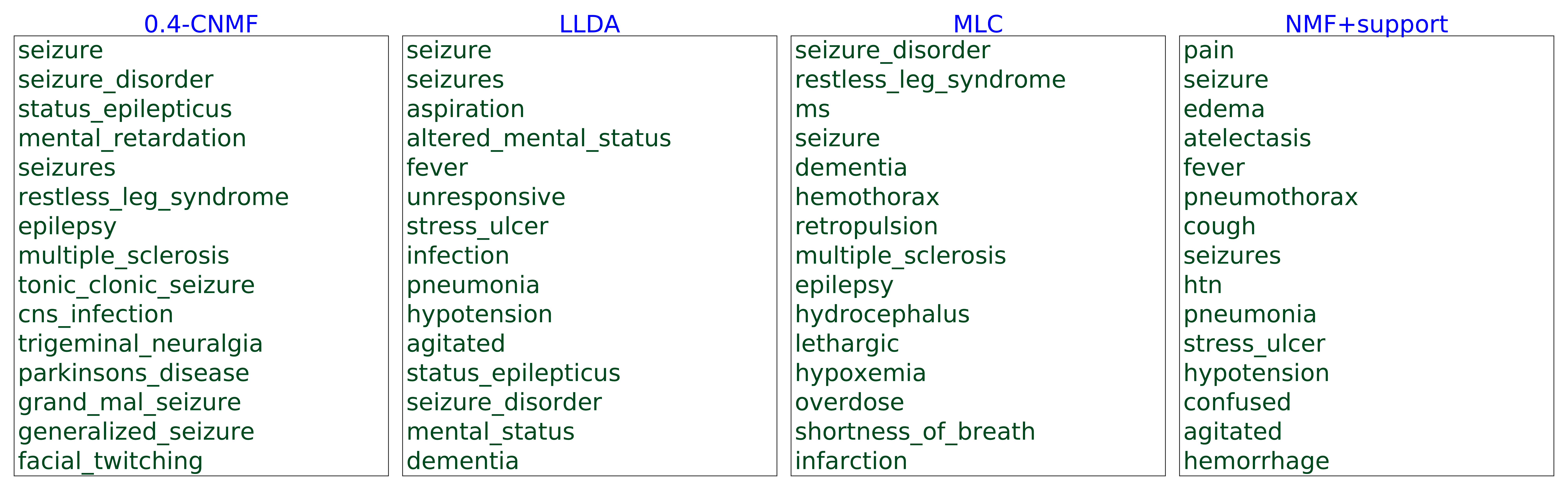}
  \caption{Learned Phenotypes for Other Neurological Disorders}
  \label{fig:other_neurological}
\end{figure}

\begin{figure}[!htbp]
\small
  \centering
  \includegraphics[width=\textwidth]{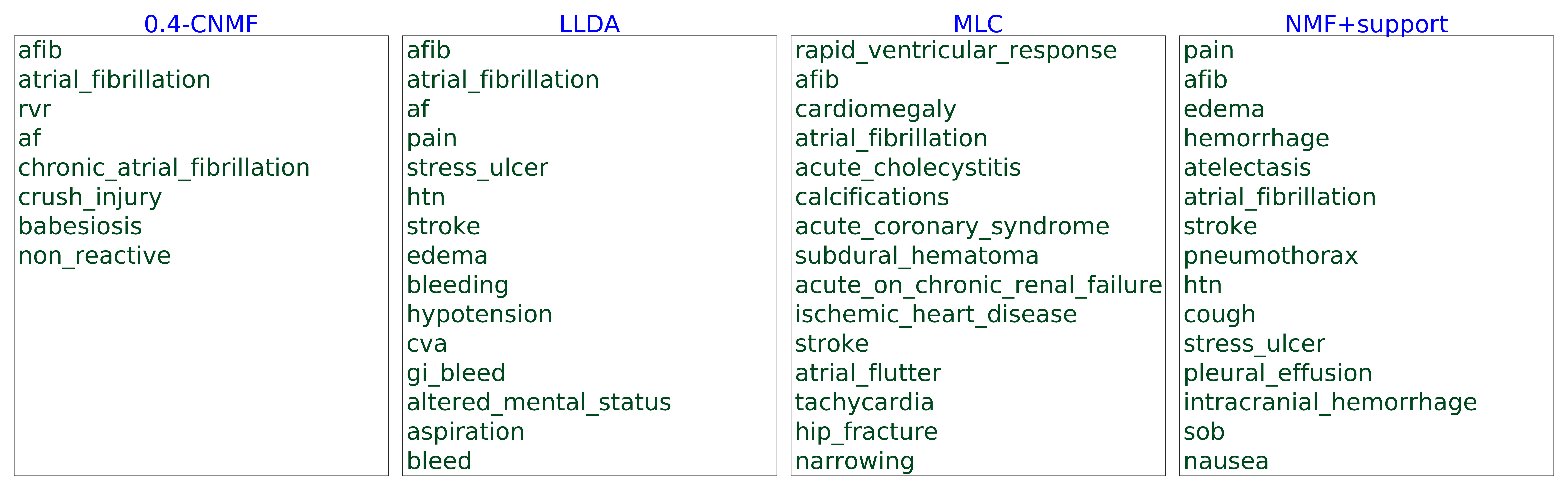}
  \caption{Learned Phenotypes for Cardiac Arrhythmias}
  \label{fig:cardiac_arrhythmias}
\end{figure}

\begin{figure}[!htbp]
\small
  \centering
  \includegraphics[width=\textwidth]{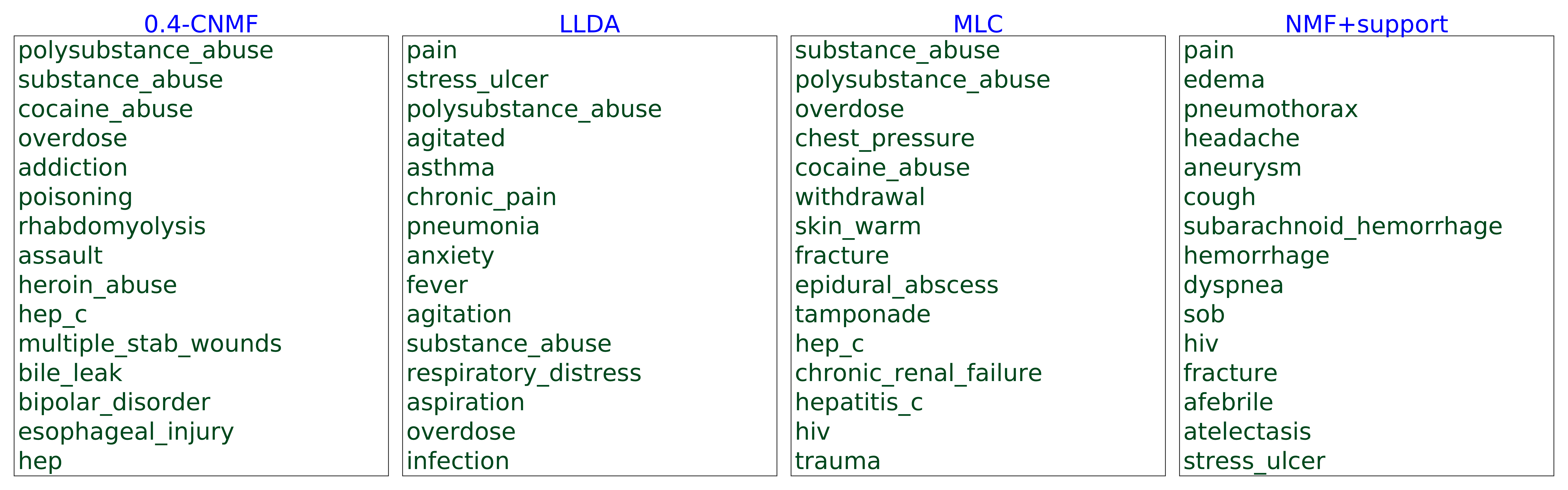}
  \caption{Learned Phenotypes for Drug Abuse}
  \label{fig:drug_abuse}
\end{figure}

\begin{figure}[!htbp]
\small
  \centering
  \includegraphics[width=\textwidth]{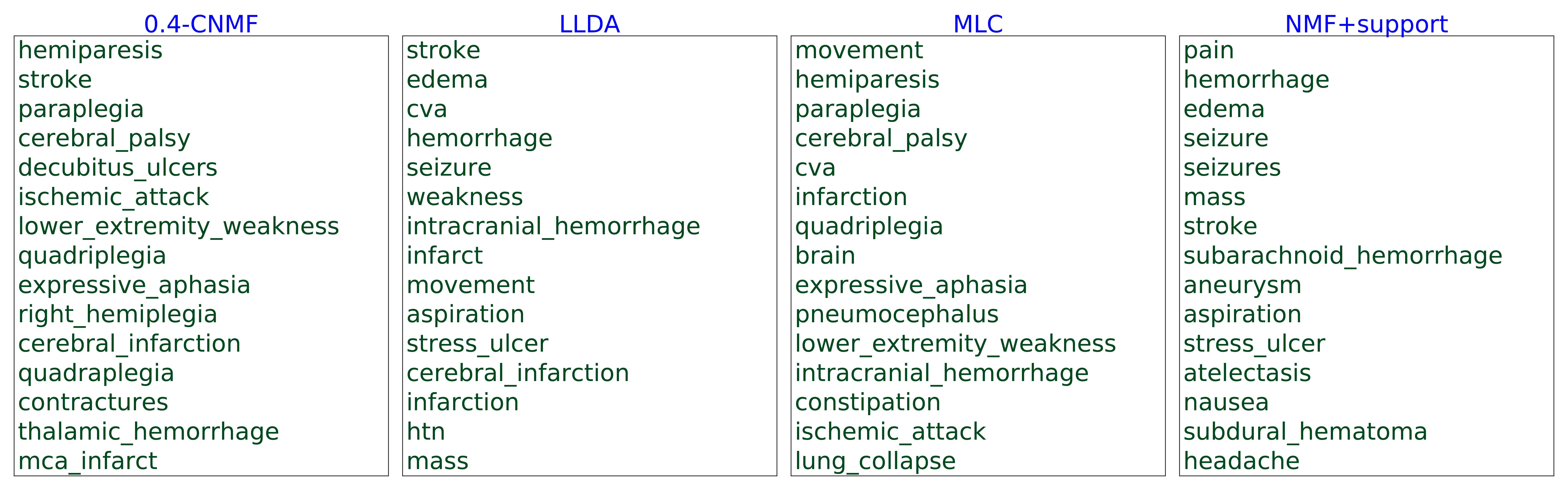}
  \caption{Learned Phenotypes for Paralysis}
  \label{fig:paralysis}
\end{figure}

\begin{figure}[!htbp]
\small
  \centering
  \includegraphics[width=\textwidth]{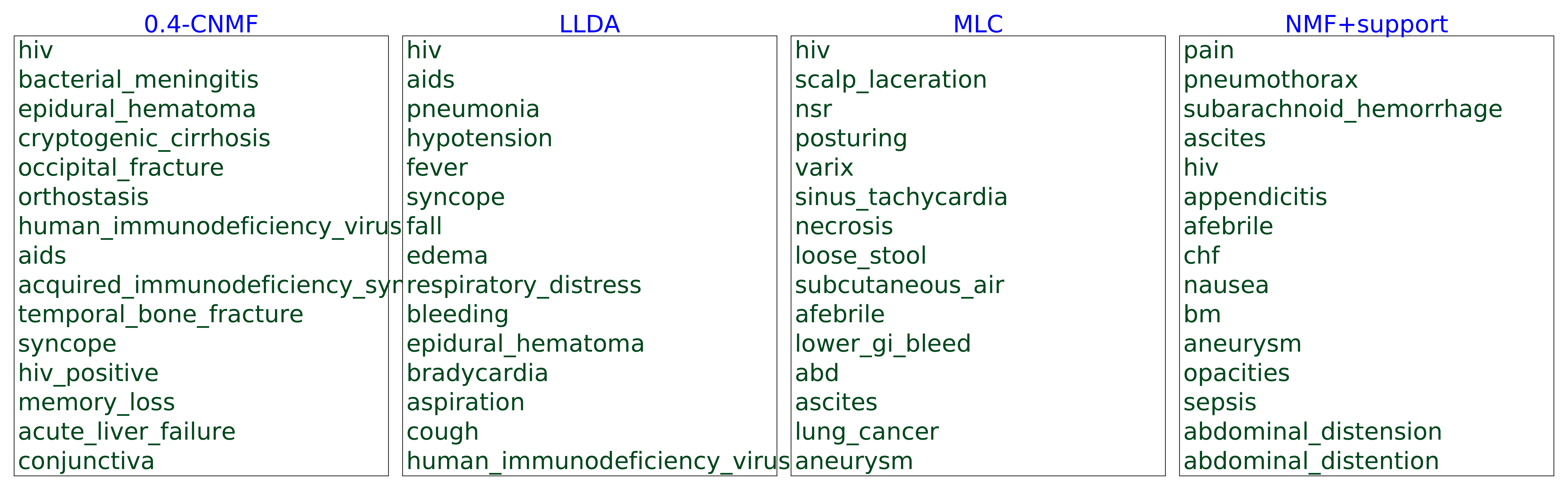}
  \caption{Learned Phenotypes for AIDS}
  \label{fig:aids}
\end{figure}

\begin{figure}[!htbp]
\small
  \centering
  \includegraphics[width=\textwidth]{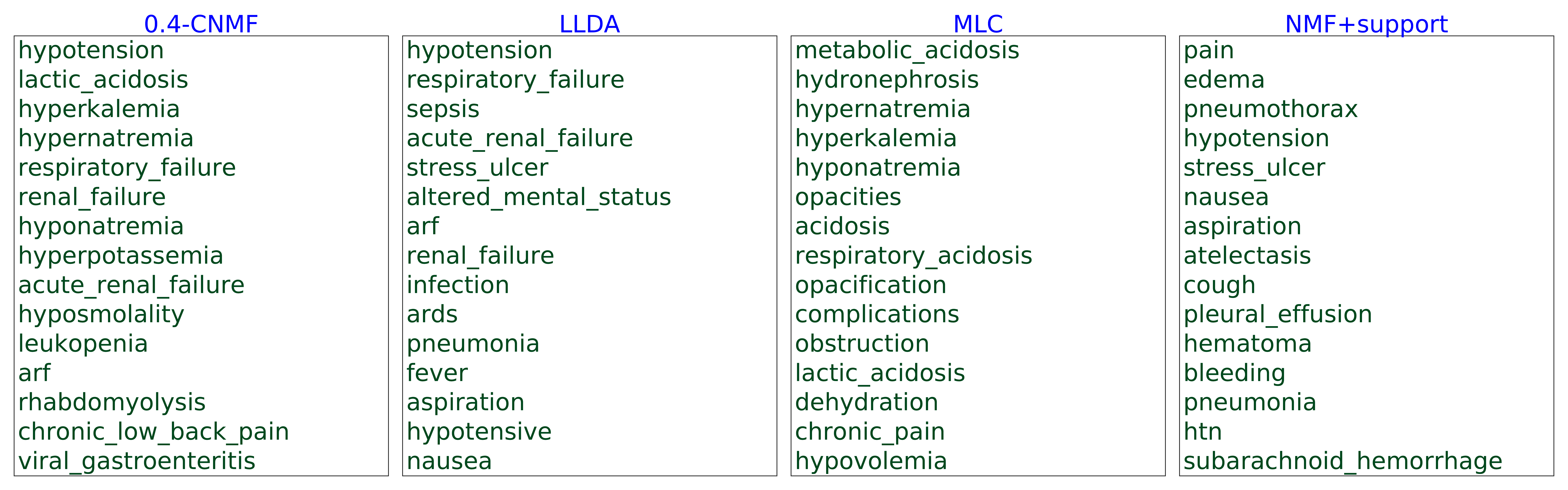}
  \caption{Learned Phenotypes for Fluid Electrolyte Disorders}
  \label{fig:fluid_electrolyte}
\end{figure}

\begin{figure}[!htbp]
\small
  \centering
  \includegraphics[width=\textwidth]{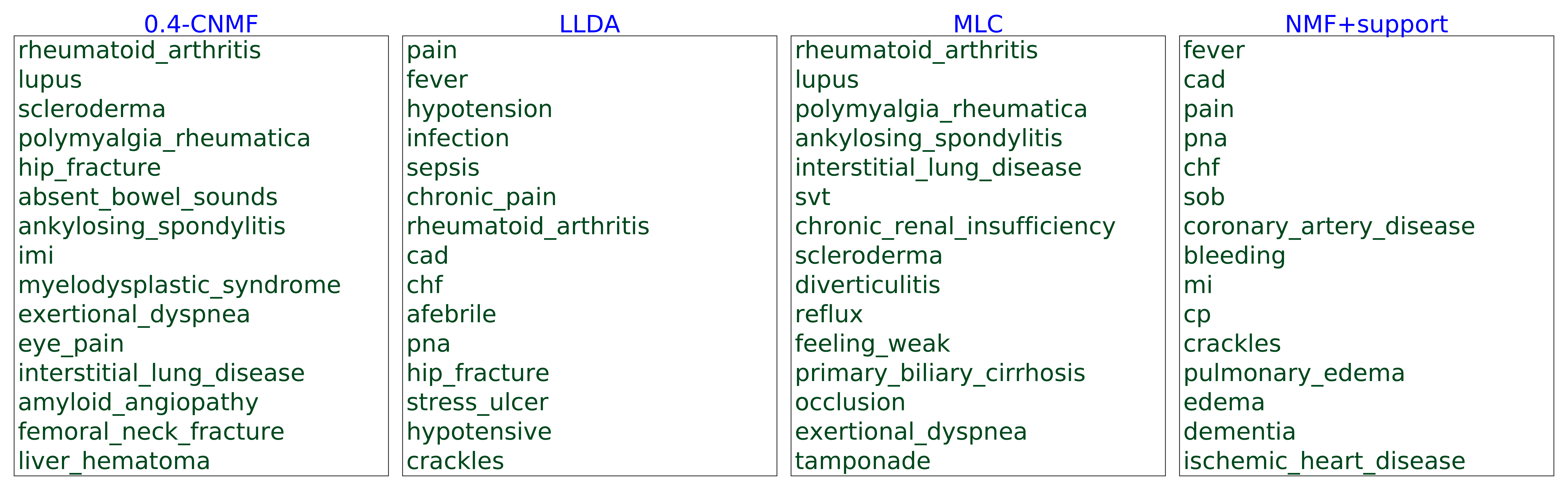}
  \caption{Learned Phenotypes for Rheumatoid Arthritis}
  \label{fig:rheumatoid_arthritis}
\end{figure}

\begin{figure}[!htbp]
\small
  \centering
  \includegraphics[width=\textwidth]{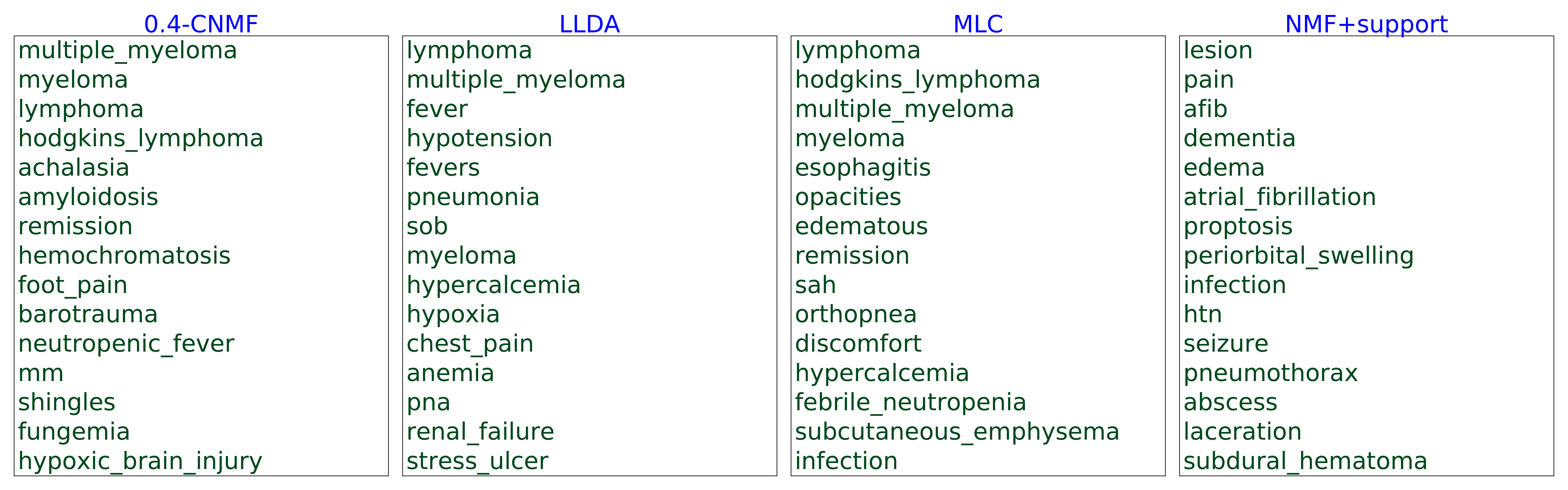}
  \caption{Learned Phenotypes for Lymphoma}
  \label{fig:lymphoma}
\end{figure}

\begin{figure}[!htbp]
\small
  \centering
  \includegraphics[width=\textwidth]{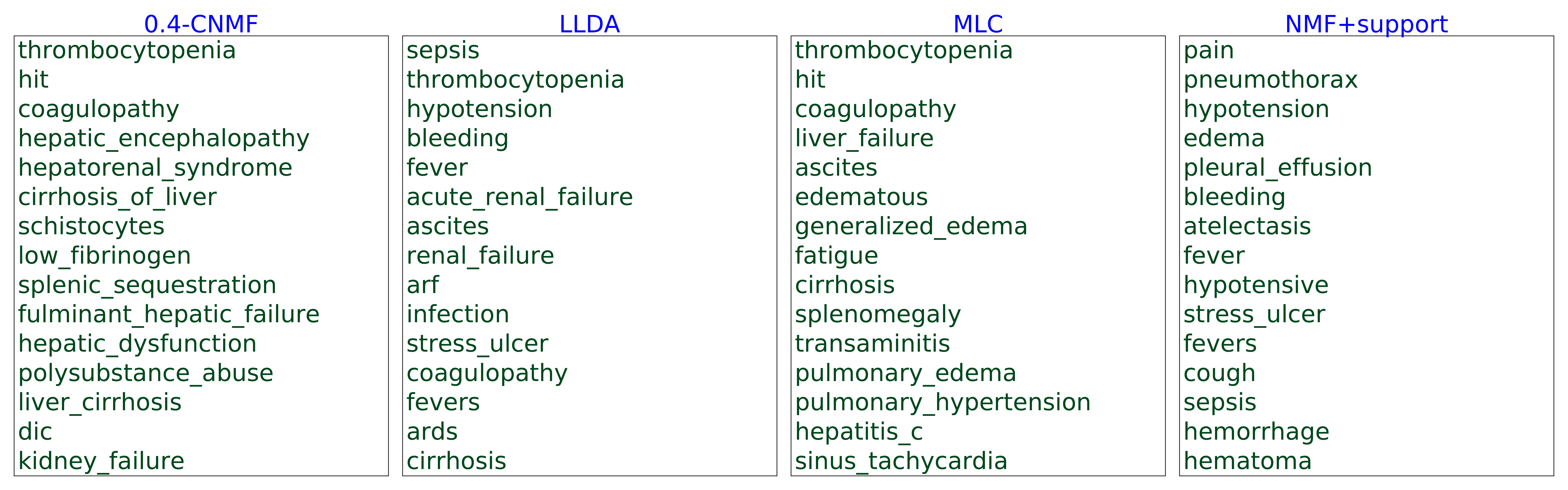}
  \caption{Learned Phenotypes for Coagulopathy}
  \label{fig:coagulopathy}
\end{figure}

\begin{figure}[!htbp]
\small
  \centering
  \includegraphics[width=\textwidth]{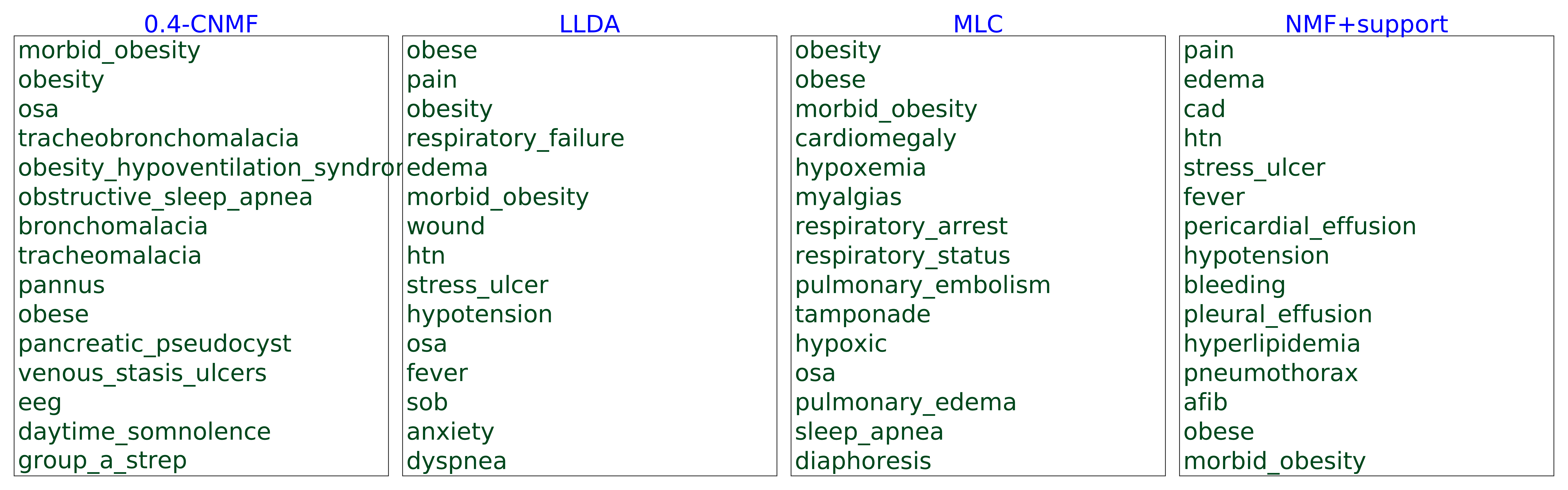}
  \caption{Learned Phenotypes for Obesity}
  \label{fig:obsesity}
\end{figure}

\begin{figure}[!htbp]
\small
  \centering
  \includegraphics[width=\textwidth]{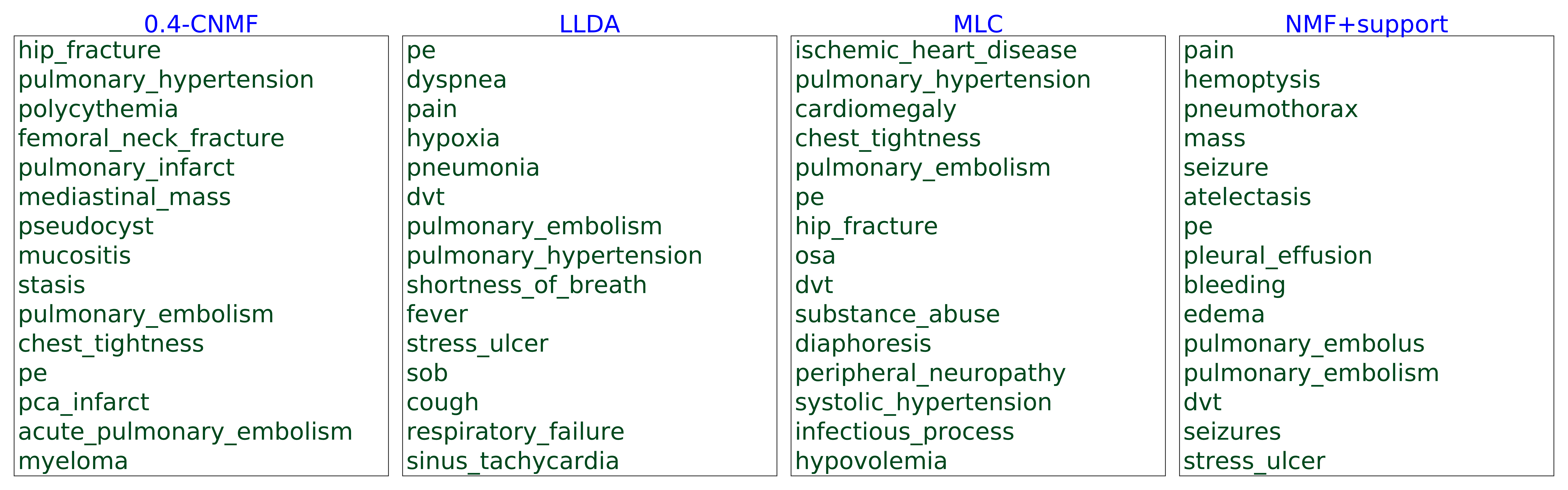}
  \caption{Learned Phenotypes for Pulmonary Circulation Disorder}
  \label{fig:pulmonary_circulation}
\end{figure}

\begin{figure}[!htbp]
\small
  \centering
  \includegraphics[width=\textwidth]{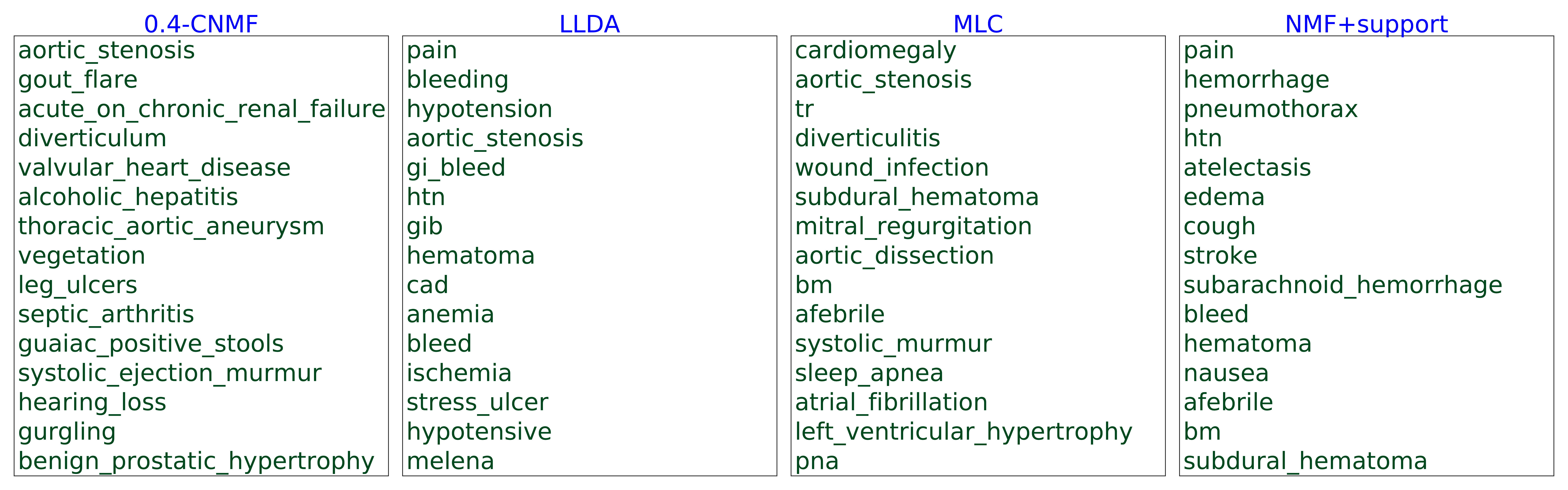}
  \caption{Learned Phenotypes for Valvular Disease}
  \label{fig:valvular_disease}
\end{figure}

\begin{figure}[!htbp]
\small
  \centering
  \includegraphics[width=\textwidth]{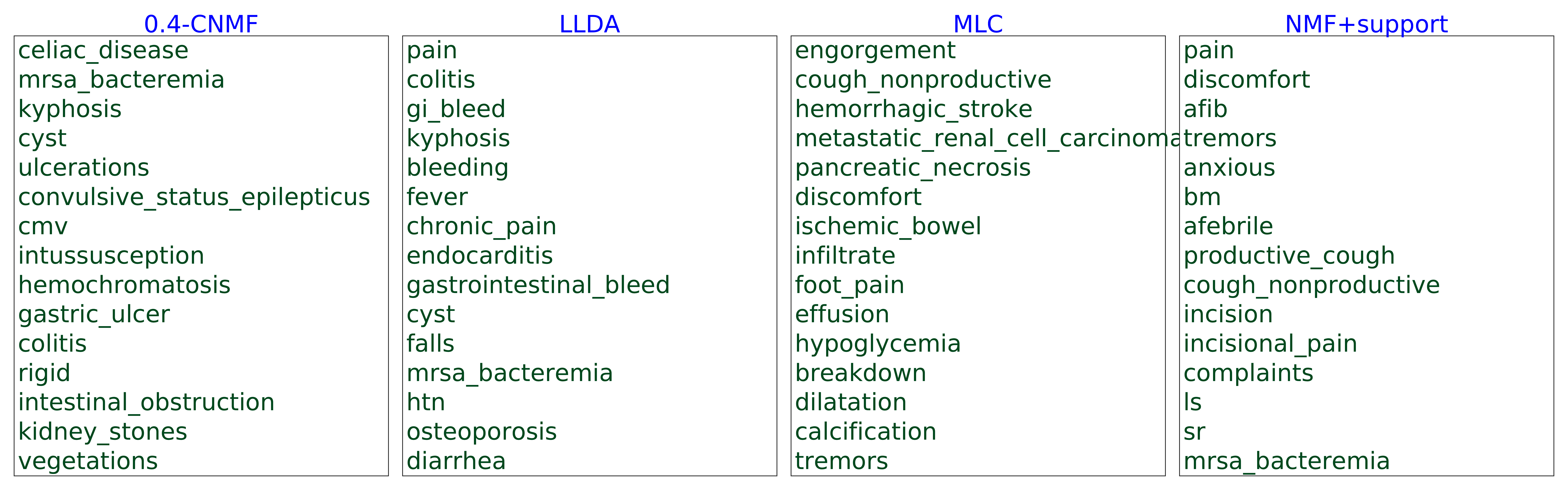}
  \caption{Learned Phenotypes for Peptic Ulcer}
  \label{fig:peptic_ulcer}
\end{figure}

\begin{figure}[!htbp]
\small
  \centering
  \includegraphics[width=\textwidth]{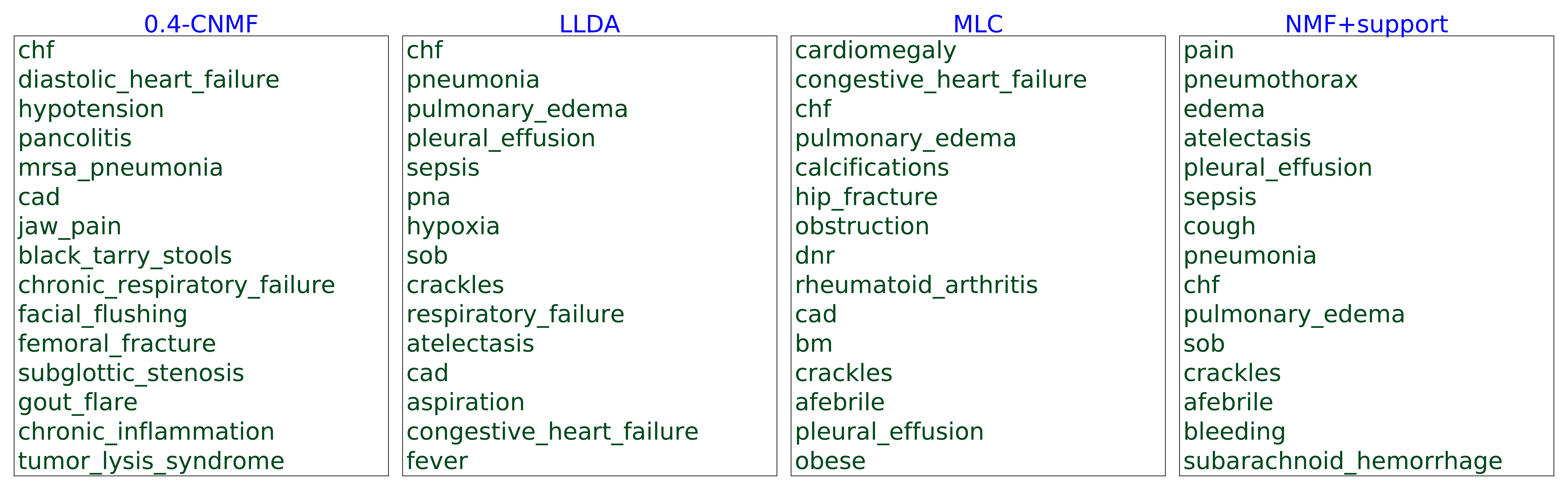}
  \caption{Learned Phenotypes for Congestive Heart Failure}
  \label{fig:congestive_heart_failure}
\end{figure}

\begin{figure}[!htbp]
\small
  \centering
  \includegraphics[width=\textwidth]{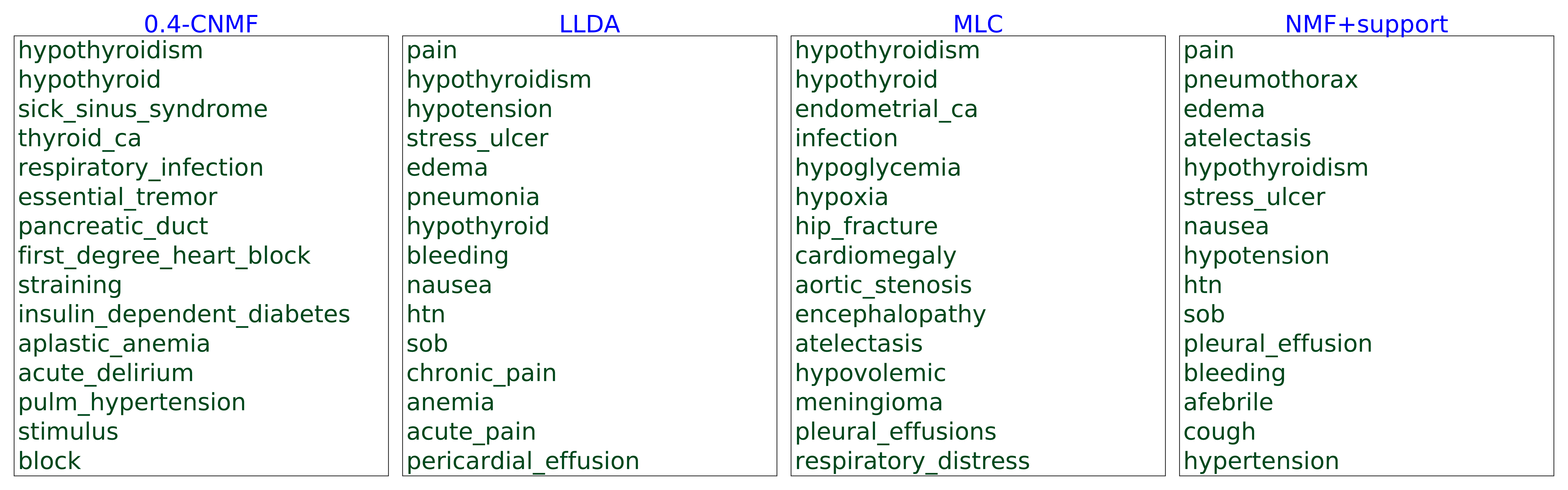}
  \caption{Learned Phenotypes for Hypothyroidism}
  \label{fig:hypothyroidism}
\end{figure}

\begin{figure}[!htbp]
\small
  \centering
  \includegraphics[width=\textwidth]{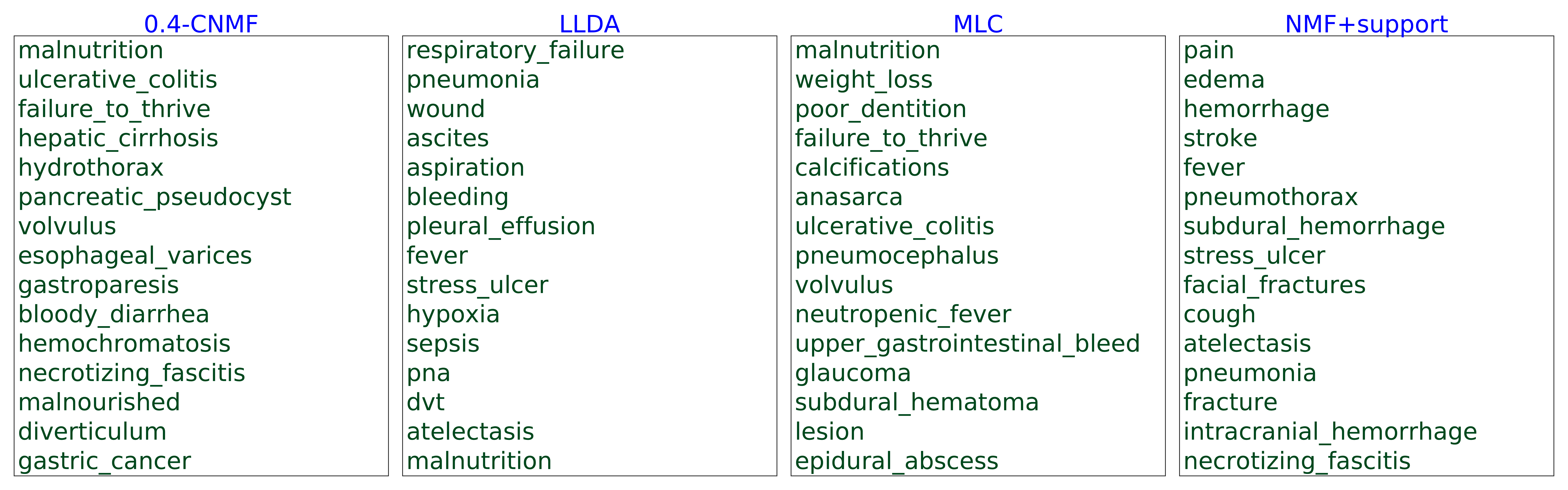}
  \caption{Learned Phenotypes for Weight loss}
  \label{fig:weight_loss}
\end{figure}

\begin{figure}[!htbp]
\small
  \centering
  \includegraphics[width=\textwidth]{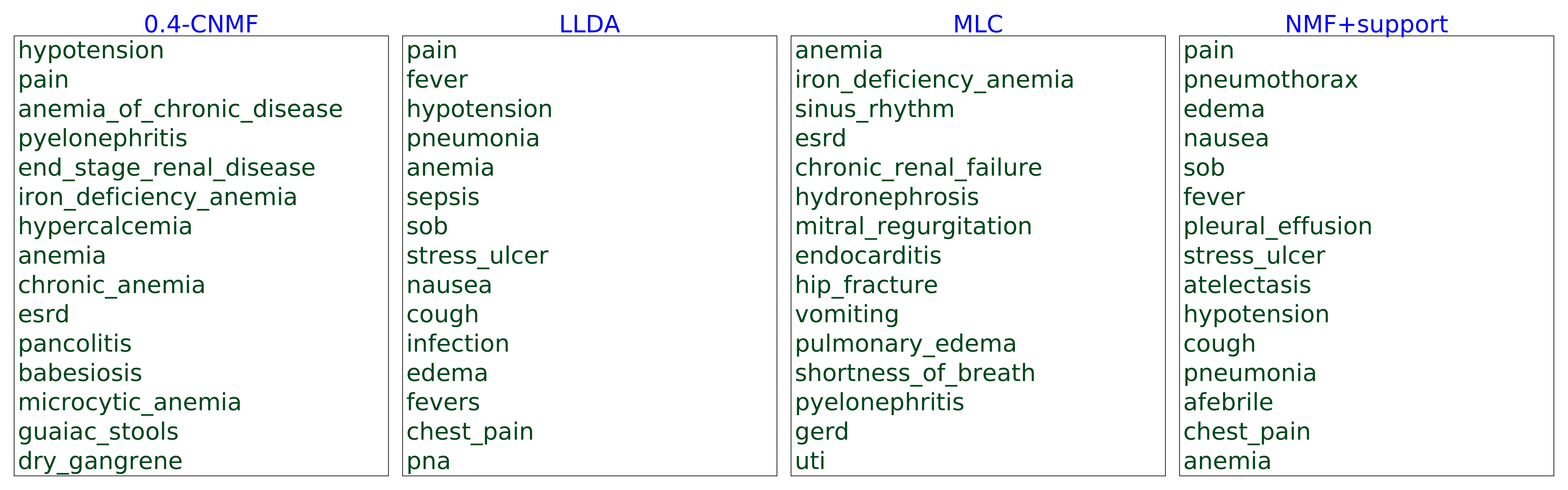}
  \caption{Learned Phenotypes for Deficiency Anemias}
  \label{fig:deficiency_anemias}
\end{figure}

\begin{figure}[!htbp]
\small
  \centering
  \includegraphics[width=\textwidth]{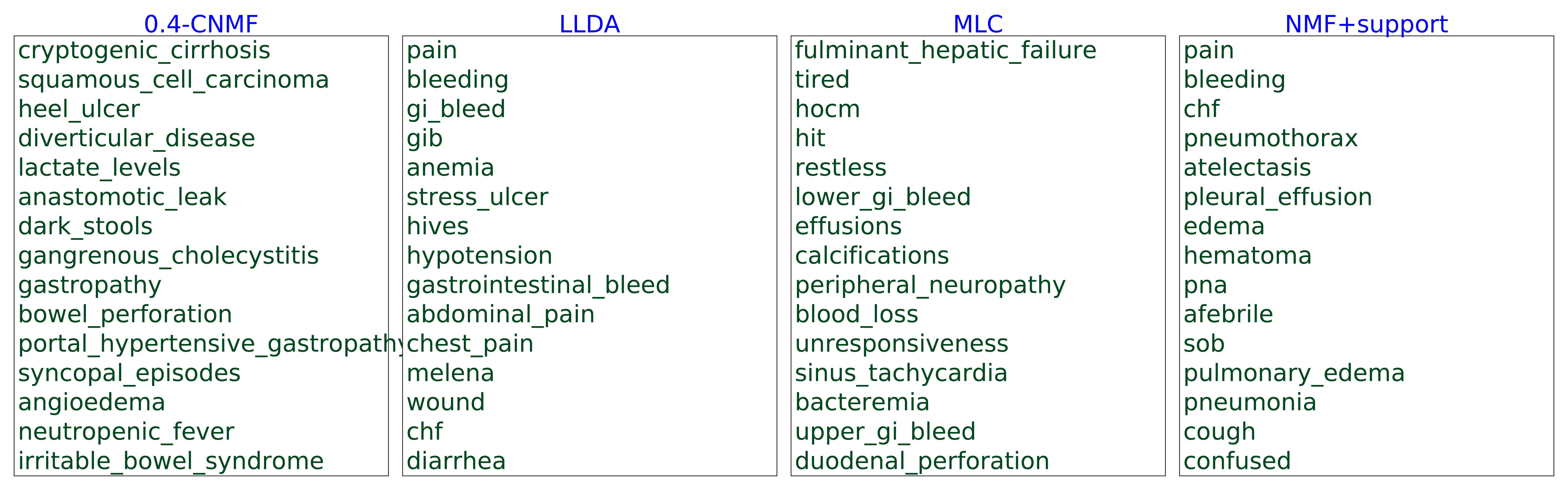}
  \caption{Learned Phenotypes for Blood Loss Anemia}
  \label{fig:blood_loss_anemia}
\end{figure}

Figures~\ref{fig:liver_disease}--\ref{fig:blood_loss_anemia} show the top $15$ terms learned for all target chronic conditions for the proposed model and baselines. %They are ordered according to the quality or ratings received for the proposed $\lambda$--CNMF baseline from both annotators. 
The sparsity level chosen is based on the criterion described in Section~\ref{sec:exp-sparsity}. For all conditions, the terms are ordered in decreasing order of importance as learned by the models.

%\begin{figure}[]
%\small
%  \centering
%  \includegraphics[width=\textwidth]{figures/phenotypes_draft/psychoses.pdf}
%  \caption{Learned Phenotypes for Psychoses}
%  \label{fig:psychoses}
%\end{figure}

\clearpage

\clearpage

%\begin{figure}[!htbp]
%\small
%  \centering
%  \includegraphics[width=\textwidth]{figures/phenotypes_draft/hypertension.pdf}
%  \caption{Learned Phenotypes for Hypertension}
%  \label{fig:hypertension}
%\end{figure}

\begin{figure}[!htbp]
\small
  \centering
  \includegraphics[width=\textwidth]{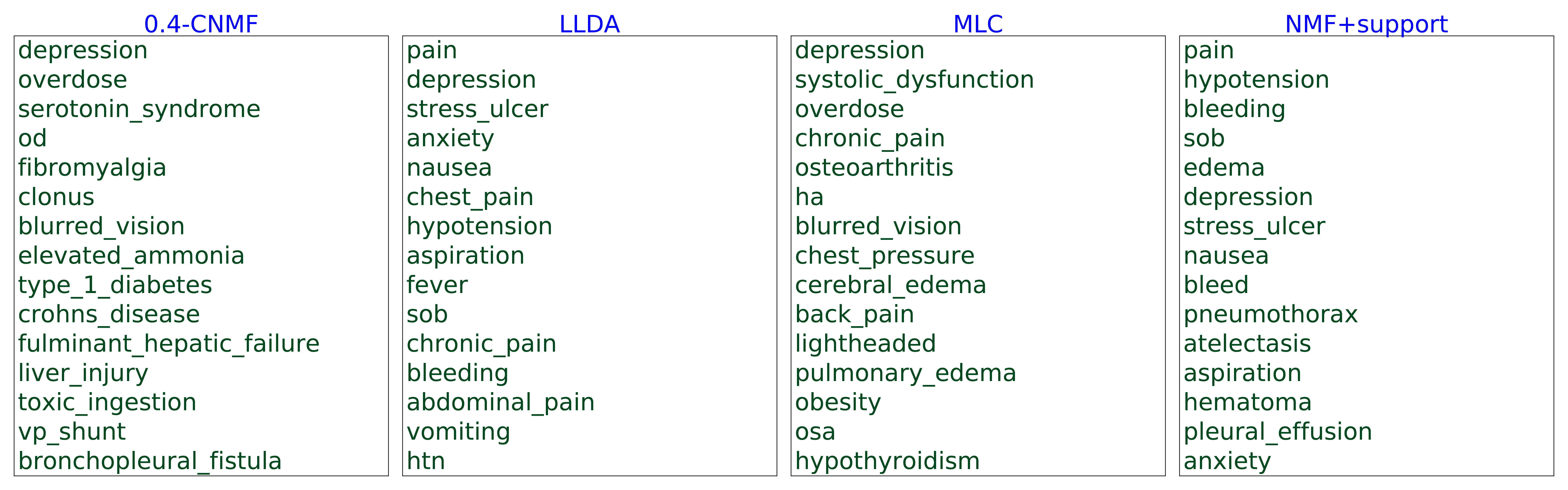}
  \caption{Learned Phenotypes for Depresssion}
  \label{fig:depression}
\end{figure}
\newpage
\section{Augmented Mortality Prediction}\label{sec:mort-pred}

\begin{figure}[htb]
\small
\centering
\includegraphics[width=0.65\textwidth]{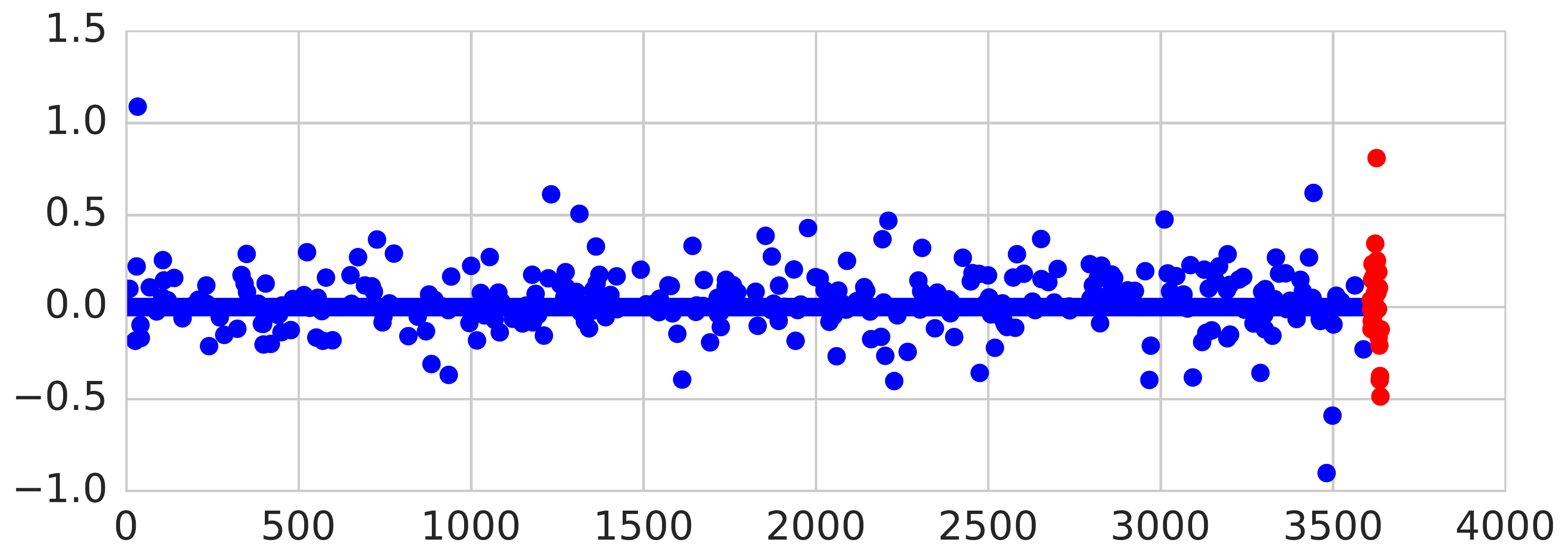}
\caption{Weights learned by the CNMF+Full EHR classifier for all features. The weights shaded red correspond to phenotypes.}% and are relatively high compared to raw notes based features, indicating that comorbidities capture significant amount of predictive information on mortality.}
\label{fig:aug_cnmf_feat1}
\end{figure}

Figure~\ref{fig:aug_cnmf_feat1} shows weights learned by the classifier for all features. The weights shaded red correspond to phenotypes and are relatively high compared to raw notes based features (shaded blue), indicating that comorbidities capture significant amount of predictive information on mortality and achieve comparable performance to full EHR model when augmented with additional raw clinical terms.

\end{document}